\documentclass[letterpaper]{article} 
\usepackage{aaai2026}  
\usepackage{times}  
\usepackage{helvet}  
\usepackage{courier}  
\usepackage[hyphens]{url}  
\usepackage{graphicx} 
\usepackage{natbib}  
\usepackage{caption} 
\usepackage[linesnumbered,ruled,vlined,noend]{algorithm2e}
\usepackage[utf8]{inputenc} 
\usepackage[T1]{fontenc}    
\usepackage{url}            
\usepackage{booktabs}       
\usepackage{amsfonts}       
\usepackage{nicefrac}       
\usepackage{microtype}      
\usepackage{xcolor}         
\usepackage{footmisc}
\usepackage{enumitem}

\usepackage{subcaption}
\usepackage{siunitx}
\usepackage{physics}
\usepackage{amssymb}
\usepackage{amsthm}
\usepackage{tikz}
\usetikzlibrary{matrix}
\usepackage{xspace}
\usepackage{multirow}

\newcommand{\VeriX}{\textsc{VeriX}}
\newcommand{\VeriXplus}{\textsc{VeriX+}}

\newcommand{\sat}{\mathtt{SAT}}
\newcommand{\unsat}{\mathtt{UNSAT}}
\newcommand{\true}{\mathtt{True}}
\newcommand{\false}{\mathtt{False}}
\newcommand{\unknown}{\mathtt{Unknown}}

\newcommand{\marabou}{\mathsf{Marabou}}

\newcommand{\network}{f}
\newcommand{\class}{c}
\newcommand{\classes}{C}
\newcommand{\image}{\mathbf{x}}
\newcommand{\pixel}{x}
\newcommand{\inDim}{m}
\newcommand{\outDim}{n}
\newcommand{\outLogits}{\mathbf{y}}
\newcommand{\outLogit}{y}
\newcommand{\varLogits}{\hat{\outLogits}}
\newcommand{\varLogit}{\hat{\outLogit}}
\newcommand{\distance}{p}
\newcommand{\magnitude}{\epsilon}
\newcommand{\discrepancy}{\delta}
\newcommand{\indexSet}{\Theta}
\newcommand{\explanation}{\mathbf{A}}
\newcommand{\irrelevant}{\mathbf{B}}

\newcommand{\getTranversalOrder}{\textsc{traversalOrder}}
\newcommand{\computeBound}{\textsc{computeBound}}
\newcommand{\binaryCompute}{\textsc{binarySequential}}

\newcommand{\lb}{\mathsf{lower}}

\newcommand{\sort}{\mathrm{sort}}
\newcommand{\checkValid}{\textsc{check}}
\newcommand{\varPixel}{\hat{\pixel}}
\newcommand{\varIn}{\hat{\image}}

\newcommand{\traversal}{\pi} 
\newcommand{\constraints}{\phi}
\newcommand{\solver}{\textsc{solve}}
\newcommand{\exitCode}{\mathsf{exitCode}}
\newcommand{\qXp}{\textsc{qXp}}
\newcommand{\QuickXplain}{\textsc{QuickXplain}}

\usepackage{amsmath}
\usepackage{amssymb}
\usepackage{mathtools}
\usepackage{amsthm}

\theoremstyle{plain}
\newtheorem{theorem}{Theorem}[section]

\theoremstyle{definition}
\newtheorem{definition}[theorem]{Definition}

\theoremstyle{remark}

\makeatletter
\newtheorem*{rep@theorem}{\rep@title}
\newcommand{\newreptheorem}[2]{%
\newenvironment{rep#1}[1]{%
 \def\rep@title{#2 \ref{##1}}%
 \begin{rep@theorem}}%
 {\end{rep@theorem}}}
\makeatother
\newreptheorem{theorem}{Theorem}
\newreptheorem{lemma}{Lemma}

\title{Efficiently Computing Compact Formal Explanations}

\author {
    Min Wu\textsuperscript{\rm 1},
    Xiaofu Li\textsuperscript{\rm 1},
    Haoze Wu\textsuperscript{\rm 2 3},
    Clark Barrett\textsuperscript{\rm 1}
}
\affiliations {
    \textsuperscript{\rm 1}Stanford University 
    \textsuperscript{\rm 2}Amherst College 
    \textsuperscript{\rm 3}VMware Research\\
    \{minwu, lixiaofu, barrettc\}@stanford.edu, hwu@amherst.edu
}

\usepackage{bibentry}

\begin{document}

\maketitle

\begin{abstract}
Building on $\VeriX$ (\textsc{veri}fied e\textsc{x}plainability), a system for producing \emph{optimal verified explanations} for machine learning models, we present $\VeriXplus$, which significantly improves both the \emph{size} and the generation \emph{time} of formal explanations. We introduce a bound propagation-based sensitivity technique to improve the size, and a binary search-based traversal with confidence ranking for improving time---the two techniques are orthogonal and can be used independently or together.  We also show how to adapt the QuickXplain algorithm to our setting to provide a trade-off between size and time. Experimental evaluations on standard benchmarks demonstrate significant improvements on both metrics, e.g., a size reduction of $38\%$ on the GTSRB dataset and a time reduction of $90\%$ on MNIST. We demonstrate that our approach is scalable to transformers and real-world scenarios such as autonomous aircraft taxiing and sentiment analysis. We conclude by showcasing several novel applications of formal explanations.
\end{abstract}

\begin{links}
    \link{Code}{github.com/NeuralNetworkVerification/VeriX+}
    \link{Extended version}{arxiv.org/abs/2409.03060}
\end{links}

\section{Introduction}
\label{sec:intro}

\emph{Explainable} AI aims to extract a set of reasoning steps from the decision-making processes of otherwise opaque AI systems, thus making them more understandable and trustworthy to humans.
Well-known work on explainable AI includes model-agnostic explainers, such as LIME~\cite{lime}, SHAP~\cite{shap}, and Anchors~\cite{anchors}, which provide explanations for neural networks by constructing a local model around a given input or identifying a subset of input features as ``anchors'' that (ideally) ensure a model's decision. While such methods can produce explanations efficiently, they do not provide \emph{formal} guarantees and thus may be inadequate in high-stakes scenarios, e.g., when transparency or fairness are paramount.

\emph{Formal} explainable AI~\cite{formalXAI} aims to compute explanations that \emph{verifiably} ensure the invariance of a model's decision. One such explanation is a minimal set of input features with the property that regardless of how the \emph{remaining} features are perturbed, the prediction remains unchanged. The intuition is that these features capture an amount of (explicit or implicit) information in the input that is sufficient to preserve the current decision. The simplest approaches allow \emph{unbounded} perturbations~\cite{abduction,PI-explanations,sufficient-reasons}, which may be overly lenient in some cases, potentially leading to explanations that are too course-grained to be useful.  As an alternative, \cite{ore} and \cite{verix} generalize these approaches by allowing both \emph{bounded} and unbounded perturbations, computing explanations with respect to such perturbations for natural language processing and perception models, respectively. The former primarily perturbs each word in a text with a finite set of its $k$ closest neighbors and thus has a discrete perturbation space; the latter considers $\magnitude$-ball perturbations over continuous and dense input spaces. The algorithm presented in \cite{verix} uses a well-known but naive approach: it simply traverses the features one by one and checks formally whether any of them can be discarded.  This approach sometimes yields overly conservative explanations that are inefficient to compute.

In this paper, we explore methods for computing better explanations by significantly improving both the \emph{size} and the generation \emph{time}. We also demonstrate novel applications that illustrate the usefulness of such explanations in practice. Our contributions can be summarized as follows.
\begin{itemize}
    \item We utilize \emph{bound propagation}-based techniques to obtain more fine-grained feature-level sensitivity information, leading to better traversal orders, which in turn produce smaller explanation sizes.
    
    \item We propose a \emph{binary search}-inspired traversal approach to enable processing features in a batch manner, thus significantly reducing the \emph{time} required to generate explanations. We also incorporate a simple but efficient \emph{confidence ranking} strategy to further reduce time.

    \item We adapt the QuickXplain algorithm~\cite{quickxplain} to provide a \emph{trade-off} between explanation \emph{size} and generation \emph{time}, and note that our adaptation is an optimization of \cite{huang2023robustness}.

    \item We demonstrate how explanations can be used in several applications, including analyzing the effects of adversarial training, as well as detecting out-of-distribution and misclassified samples. 
\end{itemize}

\section{$\VeriXplus$: \underline{Veri}fied e\underline{X}plainability plus}
\label{sec:verix+}

Let $\network$ be a neural network and $\image$ an input consisting of $\inDim$-dimensional features $\langle \pixel_1, \ldots, \pixel_\inDim \rangle$. 
The set of feature indices $\{1, \ldots, \inDim \}$ is written as $\indexSet(\image)$, or simply $\indexSet$, when the context is clear. To denote a subset of indices, we use $\explanation \subseteq \indexSet(\image)$, and $\image_\explanation$ denotes the features that are indexed by the indices in $\explanation$. We write $\network(\image) = \class$ for both regression models ($\class$ is a single quantity) and classification models ($\class \in \classes$ is one of a set of possible labels). For the latter case, we use $\outLogits = \langle \outLogit_1, \ldots, \outLogit_\outDim \rangle$ to denote confidence values for each label in $\classes$, e.g., the predicted class $\class = \arg \max (\outLogits)$. For image classification tasks, $\image$ is an image of $\inDim$ pixels, the values in $\outLogits$ represent the confidence values for each of the $\outDim$ labels in $\classes$, and $\outLogit_\class$ is the maximum value in $\outLogits$, where $\class$ is the predicted label.

\subsection{Optimal Verified Explanations}


We adopt the definition of \emph{optimal robust explanations} from \cite{ore,verix} (see also \cite{PI-explanations,abduction,sufficient-reasons}). For a network $\network$ and an input $\image$, we compute a minimal subset of $\image$, denoted by $\image_\explanation$, such that any $\magnitude$-perturbations imposed on the remaining features do not change the model's prediction.

\begin{definition}[Optimal Verified Explanation~\cite{verix}]\label{dfn:verix+}
    Given a neural network $\network$, an input $\image$, a manipulation magnitude $\magnitude$, and a discrepancy $\discrepancy$, a \emph{verified explanation} with respect to norm $\distance \in \{1, 2, \infty \}$ is a set of input features $\image_\explanation$ such that if $\irrelevant = \indexSet(\image) \setminus \explanation$, then
    \begin{equation}\label{eqn:soundness}
        \forall \ \image_\irrelevant'. \ \norm{\image_\irrelevant - \image_\irrelevant'}_\distance \leq \magnitude \implies \abs{\network(\image) - \network(\image')} \leq \discrepancy,
    \end{equation}
    where $\image_\irrelevant'$ is some perturbation on the \emph{irrelevant} features $\image_\irrelevant$ and $\image'$ is the input variant combining $\image_\explanation$ and $\image_\irrelevant'$. We say that the verified explanation $\image_\explanation$ is \emph{optimal} if
    \begin{multline}\label{eqn:optimality}
        \forall \ \pixel \in \image_\explanation. \ \exists \ \pixel', \image_\irrelevant'. \ \norm{(\pixel \cup \image_\irrelevant) - (\pixel' \cup \image_\irrelevant')}_\distance \\
        \leq \magnitude \ \wedge \abs{\network(\image) - \network(\image')} > \discrepancy,
    \end{multline}
    where $\pixel'$ is some perturbation of $\pixel \in \image_\explanation$ and $\cup$ denotes concatenation of features.
\end{definition}

Such explanations are both \emph{sound} and \emph{optimal} by Equations~(\ref{eqn:soundness}) and (\ref{eqn:optimality}), respectively~\cite{verix}.
%
%
%
Note that the optimality we define here is \emph{local} as it computes a minimal (not minimum) subset. The minimum subset is called a \emph{global} optimum, also known as the cardinality-minimal explanation~\cite{abduction}.  Finding the global optimum is often too computationally expensive to be practically useful, as it requires searching over an exponential number of local optima.

\subsection{Workflow of $\VeriXplus$ in A Nutshell}

\begin{figure}[t]
    \centering
    \includegraphics[width=\linewidth]{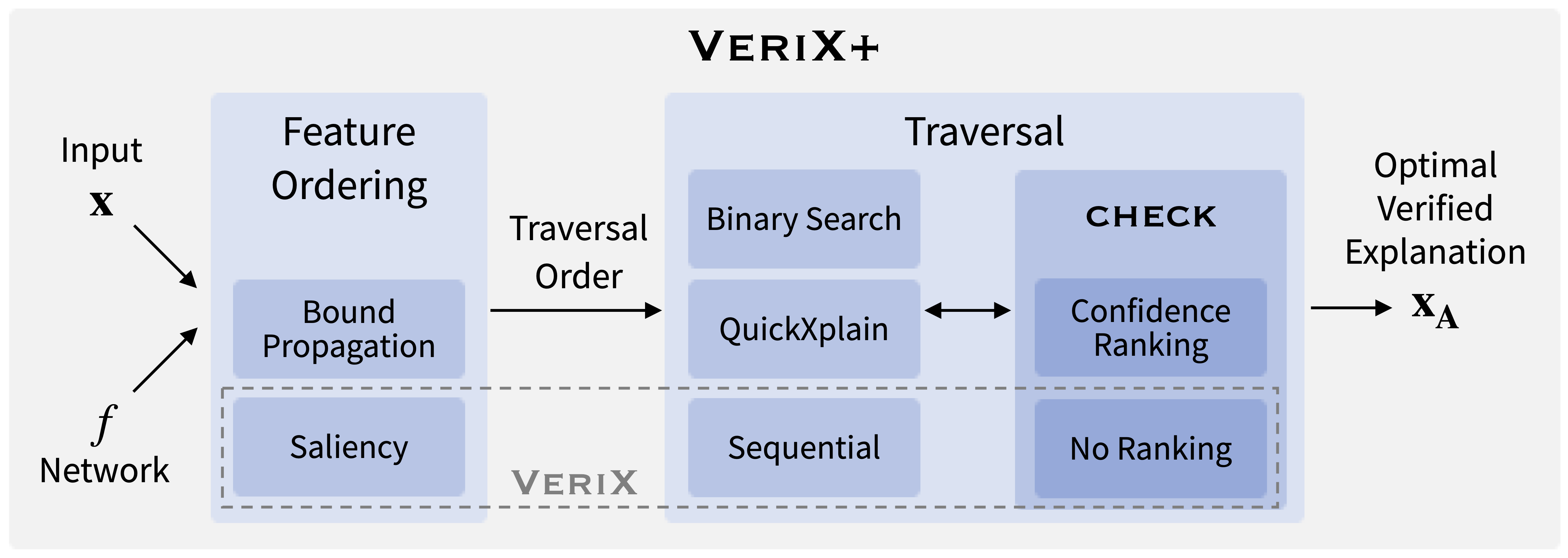}
    \caption{The $\VeriXplus$ framework.}
    \label{fig:verix+}
\end{figure}

We present the overall workflow of our $\VeriXplus$ framework in Figure~\ref{fig:verix+}. 
Starting from the left, the inputs are a network $\network$ and an input $\image$.  The first step is to obtain a sensitivity map of all the input features and, by ranking their individual sensitivity, produce a traversal order. We introduce a new bound propagation-based technique (Algorithm~\ref{alg:traversalOrder}) for obtaining more meaningful sensitivity maps and thus better traversal orders, which, in turn, reduce explanation sizes. 

The traversal order is then passed to the main traversal algorithm, which computes optimal verified explanations. We propose two new optimizations, one based on binary search (Algorithm~\ref{alg:binaryComputation}) and one adapted from the well-known QuickXplain algorithm~\cite{quickxplain} (Algorithm~\ref{alg:QuickXplain}).  
Comparing to the existing sequential method, which precesses features one at a time, the two new algorithms significantly reduce computation time by processing features in \emph{batches}. The key difference is that Algorithm~\ref{alg:binaryComputation} reduces computation time without affecting the explanation size, as it retains the original traversal order. In contrast, Algorithm~\ref{alg:QuickXplain} not only reduces computation time but also reduces the explanation size by adaptively adjusting the traversal order on the fly. This adaptive strategy introduces some overhead compared to Algorithm~\ref{alg:binaryComputation}, but it remains significantly faster than the sequential method, resulting in a trade-off between explanation size and computation time.

The $\checkValid$ procedure (Algorithm~\ref{alg:checkValid}) is used by the traversal methods to formally check the soundness of a candidate explanation. We also add a simple but efficient confidence ranking algorithm which further reduces generation time.  The confidence ranking is orthogonal to the other optimizations and benefits all three traversal approaches.


\section{Methodological Advances for Explanation Size and Generation Time}
\label{sec:methodology}

In this section, we discuss in detail several optimizations for improving both the explanation \emph{size} and the generation \emph{time}.
The bound propagation-based sensitivity ranking discussed in Section~\ref{sec:epsBounds} improves size, and the binary search-based traversal discussed in Section~\ref{sec:binarySearch} improves time, as does the confidence ranking discussed in Section~\ref{sec:conRanking}. Section~\ref{sec:QuickXplain} discusses how an adaptation of the QuickXplain algorithm~\cite{quickxplain} enables an additional size-time trade-off.

\subsection{Improving size: A Bound Propagation-based Traversal Order}
\label{sec:epsBounds}

In previous work \cite{verix}, a traversal order is computed based on a heuristic that measures how sensitive a model's confidence is to each individual feature $\pixel_i$ by imposing simple feature \textit{deletion} or \textit{reversal}.
For example, for images, after normalizing pixel values from $[0, 255]$ to $[0, 1]$, deletion of feature $i$ sets $\pixel_i=0$ and reversal sets $x_i$ to $1-\pixel_i$.
Although this produces a reasonable ranking of the input indices, it is not ideal for explanations are based on $\magnitude$-perturbations. We show how utilizing the perturbation information at the sensitivity phase can produce better traversal orders.

Procedure $\getTranversalOrder$ in Algorithm~\ref{alg:traversalOrder} computes \emph{$\magnitude$-bounds} for each feature of the input and then ranks these bounds to obtain an order on feature indices. We introduce variables $\varIn$ which represent symbolic inputs to $\network$ (Line~\ref{line:inputVar}). Then, for\footnote{The $\mathsf{for}$-loop is only used to present the functionality in a clear way; in our implementation, the $\magnitude$-bounds for all the input features are computed in \emph{parallel} for time efficiency.} each individual feature $\pixel_i$ in $\image$, we set a constraint on its corresponding variable $\varPixel_i$, requiring it to be in the \emph{interval} $[\pixel_i -\magnitude, \pixel_i + \magnitude]$ (Line~\ref{line:inputPerturb}).  Each of the remaining feature variables $\varPixel_j$, with $j \neq i$ and $j \in \{1, \ldots, \inDim \}$, is constrained to be equal to its corresponding input $\pixel_j$ (Line~\ref{line:inputFix}).\footnote{As a further optimization (not shown), if the input data are known to be bounded as part of the problem definition, we intersect the interval $[\pixel_i -\magnitude, \pixel_i + \magnitude]$ with the known bound.}
In Line~\ref{line:computeBounds}, we pass $\varIn$ (with only its $i$-th feature $\varPixel_i$ allowed to change) and the model $\network$ to a bounds-analysis procedure, which computes lower and upper bounds on the outputs of $\network (\varIn)$.  

Note that, since $\network$ could be highly nonlinear, computing these bounds is not straightforward.  That is why, instead of performing simple model inferences, we utilize existing analyses such as $\textsc{ibp}$ (interval bounded propagation)~\cite{ibp} and $\textsc{crown}$~\cite{crown} to compute the output bounds. We found (empirically) that retaining only the lower bound of the predicted class $\lb_\class$ provides the most effective ranking.  We store the lower bound in Line~\ref{line:addBound}. Once we have all the bounds, we sort them in a descending order (Line~\ref{line:sortBounds}) -- as input features producing higher lower bounds are more likely to be irrelevant to the explanation.  Intuitively, the higher the lower bound, the less the output is affected by perturbing the input. 
 Traversing less relevant features first leads to smaller explanations.

\newcommand{\newVars}{\ensuremath{\mathit{newVars()}\xspace}}

\begin{algorithm}[t]
\DontPrintSemicolon
\SetKwInOut{Parameter}{Parameter}
\SetKwProg{function}{function}{}{}
    \caption{$\getTranversalOrder$}
    \label{alg:traversalOrder}
    \KwIn{neural network $\network$ and input $\image$}
    \Parameter{$\magnitude$-perturbation}
    \KwOut{traversal order $\traversal$}
    \function{$\getTranversalOrder(\network, \image)$}{
        $\class \mapsto \network(\image)$ \; \label{line:prediction}
        $\varIn \mapsto \newVars$  \; \label{line:inputVar}
        \For{$\pixel_i \in \image$}{
            $\varPixel_i \mapsto [\pixel_i - \magnitude, \pixel_i + \magnitude]$ \; \label{line:inputPerturb}
            $\varPixel_j \mapsto [\pixel_j, \pixel_j]$ where $j \neq i$ \; \label{line:inputFix}
            $\lb \mapsto \computeBound(\network, \varIn)$ \; \label{line:computeBounds}
            $\magnitude$-bounds[i] $\mapsto \lb_\class$ \; \label{line:addBound}
        }
        $\traversal \mapsto \arg \sort (\magnitude$-bounds$, \mathsf{descending})$ \; \label{line:sortBounds}
        \Return{$\traversal$}
    }
\end{algorithm}

\subsection{Improving Time: A Binary Search-based Traversal}
\label{sec:binarySearch}

\begin{algorithm}[t]
\DontPrintSemicolon
\SetKwInOut{Parameter}{Parameter}
\SetKwProg{function}{function}{}{}
    \caption{$\binaryCompute$}
    \label{alg:binaryComputation}
    \KwIn{neural network $\network$ and input $\image$}
    \Parameter{$\magnitude$-perturbation, norm $\distance$}
    \KwOut{explanation $\image_\explanation$ and irrelevant set $\image_\irrelevant$}
    $\image_\explanation, \image_\irrelevant \mapsto \emptyset, \emptyset$ \; \label{line:initial}
    $\image_\explanation, \image_\irrelevant \mapsto \binaryCompute(\network, \image)$ \; \label{line:exeBinary}
    \function{$\binaryCompute (\network, \image_\indexSet)$}{
        \If{$\abs{\image_\indexSet}=1$ \label{line:singleFeature}}{
            \eIf{$\checkValid(\network, \image, \image_\irrelevant \cup \image_\indexSet)$ \label{line:singleCheck}}{
                $\image_\irrelevant \mapsto \image_\irrelevant \cup \image_\indexSet$ \; \label{line:singleIrr}
                \Return{}}{
            $\image_\explanation \mapsto \image_\explanation \cup \image_\indexSet$ \; \label{line:singleExp}
            \Return{} \label{line:singleReturn}}
        }
        $\image_\Phi, \image_\Psi = \mathrm{split}(\image_\indexSet, 2)$ \; \label{line:split}
        \eIf{$\checkValid(\network, \image, \image_\irrelevant \cup \image_\Phi)$ \label{line:phiCheck}}{
            $\image_\irrelevant \mapsto \image_\irrelevant \cup \image_\Phi$ \; \label{line:phiIrr}
            \eIf{$\checkValid(\network, \image,\image_\irrelevant \cup \image_\Psi)$ \label{line:psiCheck}}{
                $\image_\irrelevant \mapsto \image_\irrelevant \cup \image_\Psi$ \; \label{line:psiIrr}}{
                \eIf{$\abs{\image_\Psi}=1$ \label{line:psiSize}}{
                    $\image_\explanation \mapsto \image_\explanation \cup \image_\Psi$ \; \label{line:phiExp}}{
                    $\binaryCompute(\network, \image_\Psi)$ \; \label{line:psiBinary}}}}{\label{line:phiFalse}
        \eIf{$\abs{\image_\Phi}=1$}{
            $\image_\explanation \mapsto \image_\explanation \cup \image_\Phi$ \;}{
            $\binaryCompute(\network, \image_\Phi)$ \; \label{line:phiBinary}}
        $\binaryCompute(\network, \image_\Psi)$ \; \label{line:binary}
        }
    }
\end{algorithm}
%
Given a traversal order, the algorithm of \cite{verix} simply processes the features one by one in a sequential order. Here, we propose an improvement that processes the features using an alternative approach inspired by binary search.  The new algorithm searches for \emph{batches} of \emph{consecutive} irrelevant features.  It simultaneously checks the whole batch to see whether it is irrelevant.  If so, there is no need to process the features in the batch one by one.  We note that this approach does not change the traversal order, so the computed explanation is the same as that computed by the original sequential algorithm.

Algorithm~\ref{alg:binaryComputation} globally updates the explanation set $\image_\explanation$ and the irrelevant set $\image_\irrelevant$ throughout. Initially, these two sets are initialized to $\emptyset$. After executing $\binaryCompute(\network, \image)$, the input $\image$ is partitioned into disjoint $\image_\explanation$ and $\image_\irrelevant$, i.e., $\image_\explanation \cup \image_\irrelevant = \image$. 
%
In function $\binaryCompute(\network, \image_\indexSet)$, where $\image_\indexSet$ are candidate features, the first $\mathbf{if}$ condition (Lines~\ref{line:singleFeature}-\ref{line:singleReturn}) considers the base case when there is only a single feature left in $\image_\indexSet$. If $\checkValid(\network, \image, \image_\irrelevant \cup \image_\indexSet)$ returns $\true$ (Line~\ref{line:singleCheck}), i.e., perturbing the current $\image_\irrelevant$ and $\image_\indexSet$ does not change model's decision (same $\class$ for classification or $\abs{\class - \class'} \leq \discrepancy$ for regression), then $\image_\indexSet$ is put into $\image_\irrelevant$ (Line~\ref{line:singleIrr}).  Otherwise, it is added to $\image_\explanation$ (Line~\ref{line:singleExp}).
In the non-base case, $\image_\indexSet$ has more than just one feature. For this case, we split $\image_\indexSet$ into two sets $\image_\Phi$ and $\image_\Psi$ with similar sizes 
(Line~\ref{line:split}). If $\checkValid(\network, \image, \image_\irrelevant \cup \image_\Phi)$ returns $\true$ (Line~\ref{line:phiCheck}-\ref{line:psiBinary}), then $\image_\Phi$ belongs to the irrelevant set $\image_\irrelevant$ (Line~\ref{line:phiIrr}) and the procedure continues by checking if $\image_\Psi$ is irrelevant (Line~\ref{line:psiCheck}): if $\true$, $\image_\Psi$ is also added to $\image_\irrelevant$ (Line~\ref{line:psiIrr}). Otherwise, if $\image_\Psi$ contains only one feature (Line~\ref{line:psiSize}), we know it must be part of the explanation feature set and directly add it to $\image_\explanation$ (Line~\ref{line:phiExp}).
(Note that this check avoids unnecessary execution of the very first $\mathbf{if}$ condition (Lines~\ref{line:singleFeature}-\ref{line:singleReturn}).)
If not, we recursively call $\binaryCompute(\network, \image_\Psi)$ to search for batches of consecutive irrelevant features in $\image_\Psi$ (Line~\ref{line:psiBinary}). Finally, if $\checkValid(\network, \image, \image_\irrelevant \cup \image_\Phi)$ (Line~\ref{line:phiCheck}) returns $\false$ (Lines~\ref{line:phiFalse}-\ref{line:binary}), we similarly process $\image_\Phi$. And when $\image_\Phi$ is done, we call $\binaryCompute(\network, \image_\Psi)$ to process $\image_\Psi$ (Line~\ref{line:binary}).

\begin{theorem} [Time Complexity] \label{thm:timeBinary}
    Given a neural network $\network$ and an input $\image = \langle \pixel_1, \ldots, \pixel_\inDim \rangle$ where $\inDim \geq 2$, the \emph{time complexity} of $\binaryCompute(\network, \image)$ is $2$ calls of $\checkValid$ for the \emph{best} case (all features are irrelevant) and $k_{2\inDim} = 2 \cdot k_\inDim + 1$ or $k_{2\inDim+1} = k_{\inDim+1} + k_\inDim + 1$, for which $k_2=2$ and $k_3=4$ are the base cases, calls of $\checkValid$ for the \emph{worst} case (all features are explanatory). When $\inDim = 1$, it is obvious that only one $\checkValid$ call is needed. 
    The proof is given in Appendix~\ref{app:proofs-1}.
\end{theorem}
\noindent
Despite the fact that the worst-case complexity is worse than that of the naive algorithm (which always requires $m$ calls to $\checkValid$), we show in Section~\ref{sec:experiments} that $\binaryCompute$ performs better in practice, and remark that smaller explanations lead to more prominent advantage of our algorithm.

\subsection{Improving Time: A Confidence Ranking}
\label{sec:conRanking}

The $\checkValid$ procedure checks whether a model's decision is invariant under $\magnitude$-perturbations of a specified subset of input features.  That is, it must check whether the output $\class$ is in an allowable range for regression, i.e., $\abs{\class - \class'} \leq \discrepancy$, or whether the confidence of the predicted class $\outLogit_\class$ is always the greatest among all classes, i.e., $\outLogit_\class = \max (\outLogits)$. In the latter case, this means that in the worst case, $\abs{\outLogits}-1$ separate checks are needed to ensure that $\outLogit_\class$ is the largest.  In previous work~\cite{verix}, these checks are done naively without any insight into what order should be used.
In this work, we propose \emph{ranking} these checks based on the confidence values of the corresponding classes.  We then proceed to do the checks from the \emph{most} to the \emph{least} likely classes. If a check fails, i.e., decision invariance is not guaranteed, this is typically because we can perturb the inputs to produce one of the next most likely classes.  Thus, by checking the most likely classes first, we avoid the need for additional checks if one of those checks already fails.

\begin{algorithm}[t]
\DontPrintSemicolon
\SetKwInOut{Parameter}{Parameter}
\SetKwProg{function}{function}{}{}
    \caption{$\checkValid$ with confidence ranking}
    \label{alg:checkValid}
    \KwIn{neural network $\network$, input $\image$, and candidate features $\image_\indexSet$}
    \Parameter{$\magnitude$-perturbation}
    \KwOut{$\true / \false$}
    \function{$\checkValid(\network, \image, \image_\indexSet)$}{
        $\varIn,\varLogits \mapsto \newVars$ \;
        $\class, \outLogits \mapsto \network(\image)$ \;
        $\mathsf{ranking} \mapsto \arg \sort (\outLogits, \mathsf{descending})$ \; \label{line:conRanking}
        $\constraints \mapsto \true$
    
        \For{$i \in \indexSet$ \label{line:xPerturb}}{
            $\constraints \mapsto \constraints \land (\norm{\varPixel_i - \pixel_i}_\distance \leq \magnitude)$ \; 
        }
        \For{$i \in \indexSet(\image) \setminus \indexSet$}{
            $\constraints \mapsto \constraints \land (\varPixel_i = \pixel_i)$ \; \label{line:xFix}
        }
        \For{$j \in \mathsf{ranking} \setminus \class$ in order\label{line:forStart}}{
            $\exitCode \mapsto \solver(\network, \constraints \Rightarrow \varLogit_\class < \varLogit_j)$ \; \label{line:solver} 
            \eIf{$\exitCode == \unsat$}{
                $\mathbf{continue}$
            }{$\mathbf{break}$ \label{line:forEnd}}
        }
        \Return{$\exitCode == \unsat$ \label{line:return}}
    }
\end{algorithm}

Algorithm~\ref{alg:checkValid} shows the $\checkValid$ procedure which checks whether imposing $\magnitude$-perturbations on certain features $\image_\indexSet$ of input $\image$ maintains the decision of model $\network$: if yes, it returns $\true$, meaning that these features are irrelevant. 
To start, we create variables $\varIn$ and $\varLogits$ to represent the inputs and outputs of the model and set $\outLogits$ to be the logits and $\class$ to be the predicted class. In Line~\ref{line:conRanking}, we rank the confidence values of all the classes in a descending order, prioritizing classes that are most likely.
We allow $\magnitude$-perturbation on features in $\image_\indexSet$ while fixing the others (Lines~\ref{line:xPerturb}--\ref{line:xFix}). For each class $j$ in the sorted $\mathsf{ranking}$ (excluding $\class$ as this is the predicted class to be compared with), we call $\solver$ to examine whether the specification $\constraints \Rightarrow \varLogit_\class < \varLogit_j$ holds, i.e., whether the input constraints $\constraints$ allow a prediction change with $\varLogit_\class$ smaller than $\varLogit_j$ (Line~\ref{line:solver}). 
If the $\mathsf{exitCode}$ of $\solver$ is $\unsat$, then it means the specification is unsatisfiable in the sense that $\varLogit_\class$ will always be greater than or equal to $\varLogit_j$, i.e., the prediction cannot be manipulated into class $j$. The $\mathbf{for}$ loop (Lines~\ref{line:forStart}--\ref{line:forEnd}) examines each class in $\mathsf{ranking} \setminus \class$, and if all checks are $\unsat$ (algorithm returns $\true$ in Line~\ref{line:return}), then $\varLogit_\class$ is ensured to be the greatest among $\varLogits$, i.e., prediction invariance is guaranteed. Otherwise, if $\solver$ returns $\sat$ or $\unknown$ for any $\varLogit_j$, the algorithm returns $\false$. The key insight is that $\solver$ is more likely to return $\sat$ for classes with higher confidence, and once it does, the algorithm terminates.
In practice, $\solver$ can be instantiated with off-the-shelf neural network verification tools~\cite{deeppoly,prima,marabou,bcrown,verinet,marabou2,bound_propagation,AI2,reluplex}.

\subsection{A Size-Time \emph{Trade-off}: QuickXplain}
\label{sec:QuickXplain}

In previous sections, we propose approaches to orthogonally improve explanation size and generation time; in this section, we adapt the QuickXplain algorithm~\cite{quickxplain} and optimize it~\cite{huang2023robustness} to provide an additional \emph{trade-off} between these two metrics. We remark that the QuickXplain-based approach works as an alternative to the binary search-based method, i.e., given a traversal order from the first step of our workflow, we can either use binary search or QuickXplain.  The former improves time but does not affect size, whereas the latter affects both. In practice, QuickXplain tends to produce smaller explanations at the cost of requiring more time.  Confidence ranking benefits both techniques, as it accelerates $\checkValid$.

\begin{algorithm}[t]
\DontPrintSemicolon
\SetKwInOut{Parameter}{Parameter}
\SetKwProg{function}{function}{}{}
    \caption{$\QuickXplain$}
    \label{alg:QuickXplain}
    \KwIn{neural network $\network$ and input $\image$}
    \Parameter{$\magnitude$-perturbation, norm $\distance$}
    \KwOut{explanation $\image_\explanation$ and irrelevant set $\image_\irrelevant$}
    $\image_\explanation, \image_\irrelevant \mapsto \qXp(\emptyset, \emptyset, \image)$ \; \label{line:callQuickXplain}
    \function{$\qXp (\image_\alpha, \image_\beta, \image_\indexSet)$ \label{line:QuickXplain}}{
        \If{$\abs{\image_\indexSet}=1$ \label{line:QXPsingleX}}{
            \eIf{$\checkValid(\network, \image, \image_\beta \cup \image_\indexSet)$}{ \label{line:QXPCheckSingle}
                \Return{$\emptyset, \image_\beta \cup \image_\indexSet$ \label{line:QXPsingleXirr}}}{
            \Return{$\image_\indexSet, \image_\beta$} \label{line:QXPsingleX_}}
        }
        $\image_\Phi, \image_\Psi = \mathrm{split}(\image_\indexSet, 2)$ \; \label{line:QXPsplit}
        \uIf{$\checkValid(\network, \image, \image_\beta \cup \image_\Phi)$ \label{line:QXPphiIf}}{
            \Return{$\qXp(\image_\alpha, \image_\beta \cup \image_\Phi, \image_\Psi)$ \label{line:QXPpsi}}
        }
        \uElseIf{$\checkValid(\network, \image, \image_\beta \cup \image_\Psi)$}{ \label{line:QXPpsiIf}
            \Return{$\qXp(\image_\alpha, \image_\beta \cup \image_\Psi, \image_\Phi)$ \label{line:QXPphi}}
        }
        \Else{\label{line:QXP}
            \eIf{$\abs{\image_\Phi} = 1$}{
                $\image_\Phi', \image_\beta' \mapsto \image_\Phi, \image_\beta$ \; \label{line:QXP1phi}}{
            $\image_\Phi', \image_\beta' \mapsto \qXp( \image_\alpha \cup \image_\Psi, \image_\beta, \image_\Phi)$ \label{line:newQXPphi}
            }
            \eIf{$\abs{\image_\Psi} = 1$}{
                 $\image_\Psi', \image_\beta'' \mapsto \image_\Psi, \image_\beta'$ \; \label{line:QXP1psi}}{
            $\image_\Psi', \image_\beta'' \mapsto \qXp(\image_\alpha \cup \image_\Phi', \image_\beta', \image_\Psi)$ \label{line:newQXPpsi}
            }
            \Return{$\image_\Phi' \cup \image_\Psi', \image_\beta''$ \label{line:QXPreturn}}
        }
    }
\end{algorithm}
%
%
We present our $\QuickXplain$ adaptation in Algorithm~\ref{alg:QuickXplain}. 
The function $\qXp(\image_\alpha, \image_\beta, \image_\indexSet)$ itself is recursive with three arguments: (1) the current explanation $\image_\alpha$; (2) the current irrelevant set $\image_\beta$; and (3) the current (sub)set of input features that need to be analyzed, $\image_\indexSet$. These three sets always form a partition of the full set of features. To start with, $\image_\alpha$ and $\image_\beta$ are initialized to $\emptyset$, and when $\qXp$ proceeds, irrelevant features are added into $\image_\beta$ in a monotonically increasing way; finally, after all features are done, $\image_\alpha$ is returned as the optimal explanation $\image_\explanation$ and $\image_\beta$ as the irrelevant set $\image_\irrelevant$. 
Now we walk through the algorithm. 
Lines~\ref{line:QXPsingleX}--\ref{line:QXPsingleX_} cover the base case when $\image_\indexSet$ has a single feature as in Algorithm~\ref{alg:binaryComputation}.
When there is more than one feature in $\image_\indexSet$, it is split into two subsets $\image_\Phi$ and $\image_\Psi$ (Line~\ref{line:QXPsplit}). In Lines~\ref{line:QXPphiIf}--\ref{line:QXPphi}, we check if the subset $\image_\Phi$ or $\image_\Psi$ belongs to the irrelevant set: if $\true$, then we add it into $\image_\beta$ when calling $\qXp$ to process the other subset. 
%
If neither $\image_\Phi$ nor $\image_\Psi$ is irrelevant, we take turns processing $\image_\Phi$ and $\image_\Psi$ in Lines~\ref{line:QXP}--\ref{line:QXPreturn}: the first $\mathbf{if}$ condition analyzes $\image_\Phi$, and the second $\mathbf{if}$ processes $\image_\Psi$. If either of them has only a single feature, then we know it must be included as an explanation feature ($\image_\Phi'$ on Line~\ref{line:QXP1phi} or $\image_\Psi'$ on Line~\ref{line:QXP1psi}). This avoids the unnecessary execution of  Lines~\ref{line:QXPsingleX}--\ref{line:QXPsingleX_} in a recursive call since we already know the feature cannot be irrelevant.\footnote{This enables Algorithm~\ref{alg:QuickXplain} to have \emph{fewer} calls to $\checkValid$ (i.e., oracle calls) than the similar adaptation of QuickXplain in \cite{huang2023robustness}. \label{footnote:quickXplain}}
If either of them has more than one feature, then $\qXp$ is called recursively: $\image_\Phi$ is processed in Line~\ref{line:newQXPphi}, with $\image_\Psi$ as part of the new $\image_\alpha$, and $\image_\Psi$ is processed in Line~\ref{line:newQXPpsi} with $\image_\Phi'$ as part of the new $\image_\alpha$. Finally, both $\image_\Phi'$ and $\image_\Psi'$ are returned as explanatory features, and $\image_\beta''$ as the irrelevant set (Line~\ref{line:QXPreturn}).

\begin{theorem} [Time Complexity] \label{thm:timeQuickXplain}
    Given a neural network $\network$ and an input $\image = \langle \pixel_1, \ldots, \pixel_\inDim \rangle$ where $\inDim \geq 2$, the \emph{time complexity} of $\qXp(\emptyset, \emptyset, \image)$ is $\lfloor \log_2 \inDim \rfloor + 1$ calls of $\checkValid$ in the \emph{best} case (all features are irrelevant) and $(\inDim-1) \times 2$ calls of $\checkValid$ in the \emph{worst} case (all features are explanatory). When $\inDim = 1$, it is obvious that only one $\checkValid$ call is needed. 
    The proof is given in Appendix~\ref{app:proofs-2}.
\end{theorem}

\begin{figure}[t]
    \centering
    \begin{subfigure}{0.325\linewidth}
        \includegraphics[width=0.485\linewidth]{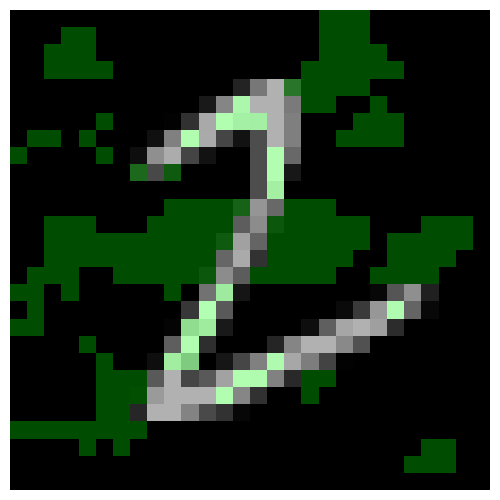}
        \includegraphics[width=0.485\linewidth]{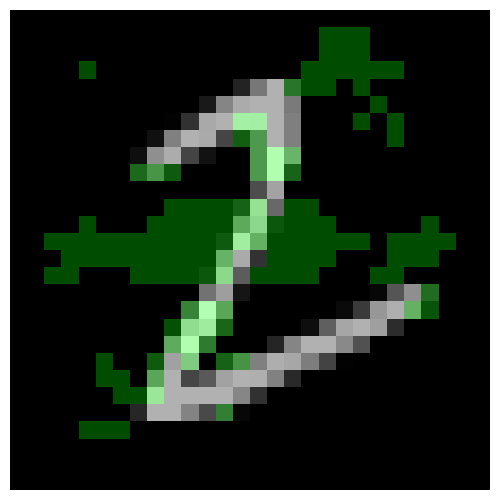}
        \caption{MNIST}
        \label{fig:verix_verix+_mnist}
    \end{subfigure}
    \begin{subfigure}{0.325\linewidth}
        \includegraphics[width=0.485\linewidth]{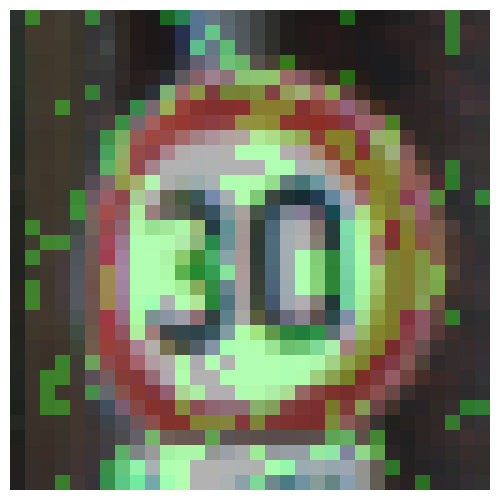}
        \includegraphics[width=0.485\linewidth]{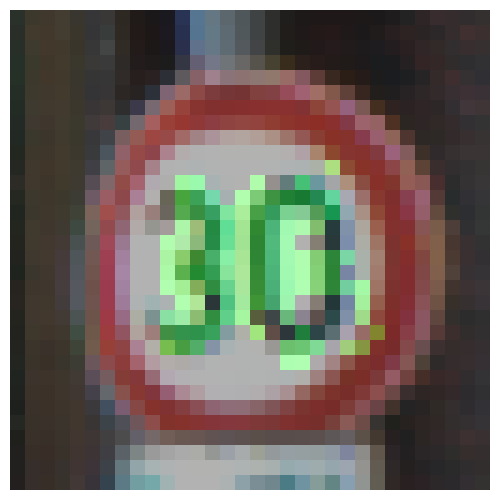}
        \caption{GTSRB}
        \label{fig:verix_verix+_gtsrb}
    \end{subfigure}
    \begin{subfigure}{0.325\linewidth}
        \includegraphics[width=0.485\linewidth]{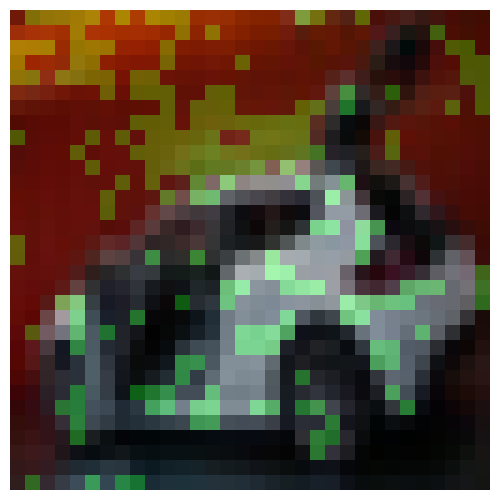}
        \includegraphics[width=0.485\linewidth]{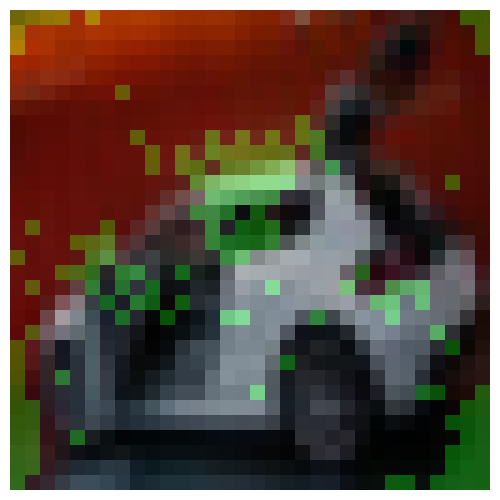}
        \caption{CIFAR10}
    \end{subfigure}
    \caption{Explanations (highlighted in green) from $\VeriX$ (left) and $\VeriXplus$ (right). (a) MNIST: size $187$ vs. $137$; time $635$ vs. $51$.  (b) GTSRB: size $276$ vs. $117$; time $703$ vs. $49$. (c) CIFAR10: size $208$ vs. $146$; time $482$ vs. $57$. Statistical significance for size and time reduction is in Table~\ref{tab:size-reduction}.}
    \label{fig:verix_verix+}
\end{figure}

\begin{figure}[t]
    \centering
    \begin{subfigure}{0.352\linewidth}
        \centering
        \includegraphics[width=\linewidth]{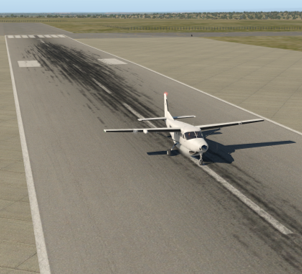}
    \end{subfigure}
    \begin{subfigure}{0.31\linewidth}
        \centering
        \begin{subfigure}{\linewidth}
            \centering
            \includegraphics[width=2.53cm]{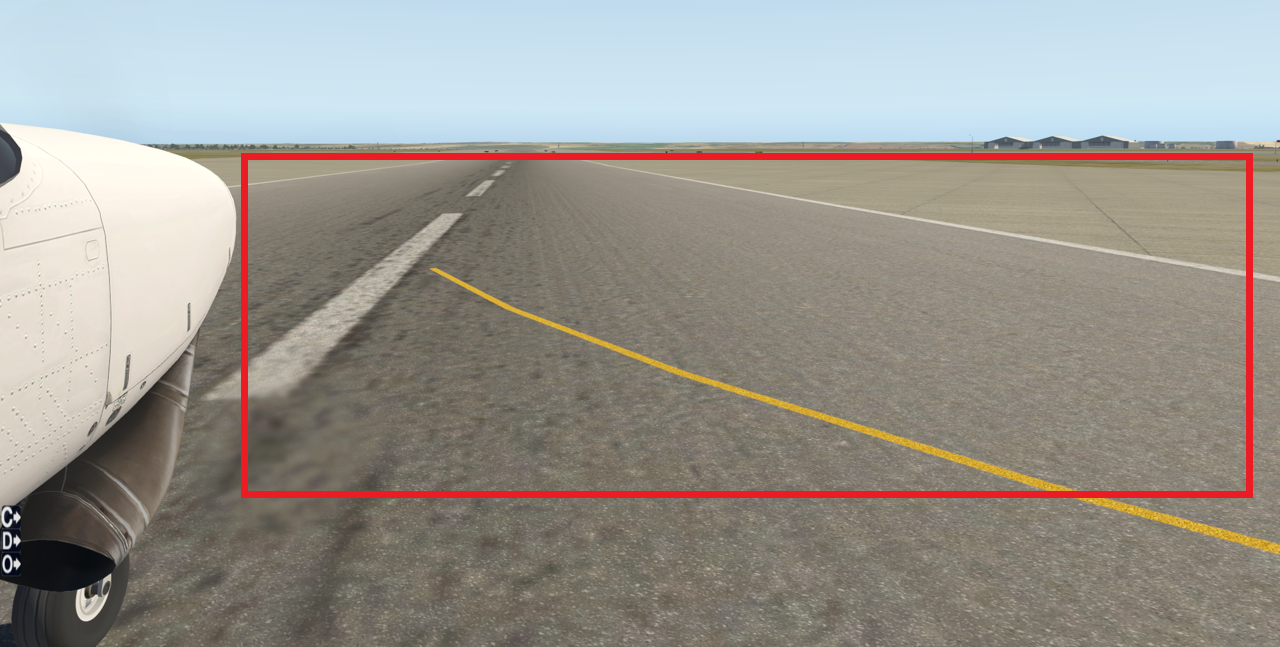}
            \vspace{-9pt}
        \end{subfigure}
        \begin{subfigure}{\linewidth}
            \centering
            \includegraphics[width=\linewidth]{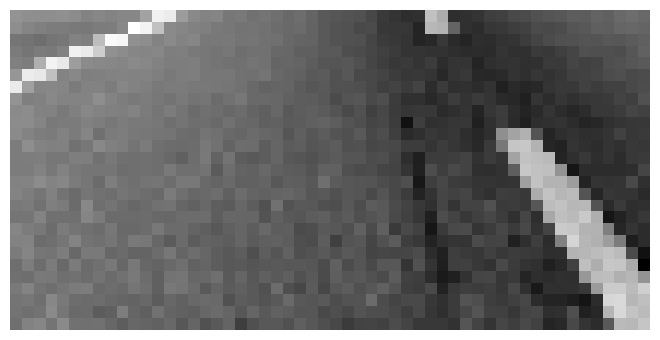}
            \vspace{-12pt}
        \end{subfigure}
        \end{subfigure}
    \begin{subfigure}{0.31\linewidth}
        \centering
        \begin{subfigure}{\linewidth}
            \centering
            \includegraphics[width=\linewidth]{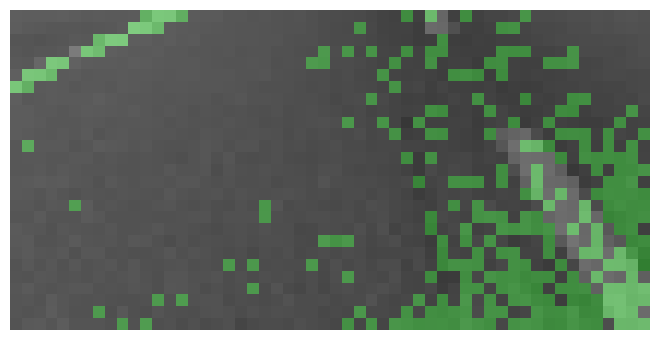}
            \vspace{-10pt}
        \end{subfigure}
        \begin{subfigure}{\linewidth}
            \centering
            \includegraphics[width=\linewidth]{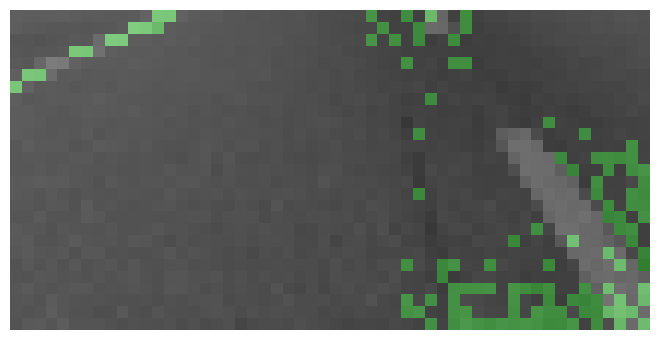}
            \vspace{-12pt}
        \end{subfigure}
    \end{subfigure}
    \caption{A real-world aircraft taxiing scenario~\cite{taxinet}. Pictures are taken from the camera fixed on the right wing of the aircraft. Explanation (green) from the transformer model (bottom right) is more compact than its fully-connected counterpart (top right).}
    \label{fig:taxiing}
\end{figure}

\section{Experimental Results}
\label{sec:experiments}

We have implemented the $\VeriXplus$ framework in Python. To realize the $\computeBound$ analysis in Algorithm~\ref{alg:traversalOrder}, we utilize the $\mathsf{bound\text{-}propagation}$\footnote{\scalebox{1}{\url{https://pypi.org/project/bound-propagation/}}} \cite{Mathiesen2022} package for fully-connected models and the $\mathsf{auto\text{\_}LiRPA}$\footnote{\scalebox{1}{\url{https://github.com/Verified-Intelligence/auto_LiRPA}}} \cite{bound_propagation} library for convolutional models. While the latter also supports dense layers, the former computes tighter $\textsc{ibp}$ bounds which, in our case, lead to smaller explanations. 
We use $\marabou$ \cite{marabou2},\footnote{\scalebox{1}{\url{https://github.com/NeuralNetworkVerification/Marabou}}} a neural network verification tool, to perform the $\solver$ function in Algorithm~\ref{alg:checkValid}. Our framework can accommodate other existing tools as long as they can perform the $\computeBound$ or the $\solver$ functionality.
We trained fully-connected and convolutional networks on standard image benchmarks
including MNIST~\cite{mnist}, GTSRB~\cite{gtsrb}, and CIFAR10~\cite{cifar10}, 
as well as transformers~\cite{attention} for real-world applications. Details on the model structures are in Appendix~\ref{app:models}. 
Experiments were performed on a workstation equipped with 16 AMD Ryzen\textsuperscript{\tiny TM} 7 5700G CPUs and 32GB memory running Fedora 37.

\begin{table*}[t]
    \caption{Average explanation \emph{size} (number of pixels) and generation \emph{time} (seconds) for both $\VeriX$ and $\VeriXplus$ on fully-connected (-FC), convolutional (-CNN), and transformer (-Transformer) models.
    $\magnitude$ is set to $5\%, 1\%$, and $1\%$ for the MNIST, GTSRB, and CIFAR10 datasets, respectively. We also include transformer results on TaxiNet and IMDB with $\magnitude=1\%$ and $0.05$. For images, we use Chebyshev distance to mimic natural distortions; for texts, we use Euclidean distance on their embeddings. 
    }
    \label{tab:verix_verix+}
    \centering
    \setlength{\tabcolsep}{4pt} 
    \renewcommand{\arraystretch}{1.2}
        \scalebox{0.9}{
    \begin{tabular}{l|cc|cc|cc|cc|cc|cc|cc}
        \toprule
        approaches & \multicolumn{2}{c|}{$\VeriX$}  & \multicolumn{12}{c}{\textbf{$\VeriXplus$}} \\ 
        traversal order & \multicolumn{2}{c|}{saliency} & \multicolumn{6}{c|}{saliency} & \multicolumn{6}{c}{\emph{bound propagation}} \\
        traversal procedure & \multicolumn{2}{c|}{sequential} & \multicolumn{2}{c|}{\emph{conf ranking}} & \multicolumn{2}{c|}{\emph{binary search}} & \multicolumn{2}{c|}{\emph{QuickXplain}} & \multicolumn{2}{c|}{\emph{conf ranking}} & \multicolumn{2}{c|}{\emph{binary search}} & \multicolumn{2}{c}{\emph{QuickXplain}} \\
        metrics & size & time & size & time & size & time & size & time  & size & time & size & time & size & time  \\ \hline \hline
        MNIST-FC & $186.7$ & $92.3$ & $186.7$ &  $81.4$ & $186.7$ &  $29.6$ & $186.3$ & $32.1$ & $179.5$ & $70.6$ &  $179.5$ &  $\mathbf{28.4}$ & $\mathbf{179.0}$ & $32.2$ \\
        MNIST-CNN &  $105.7$ &  $439.2$ & $105.7$ &  $398.9$ & $105.7$ &  $46.3$ & $105.7$ & $51.5$ & $100.8$ & $295.8$ &  $100.8$ &  $\mathbf{42.0}$ & $\mathbf{100.6}$ & $47.4$ \\
        GTSRB-FC &  $529.4$ &  $614.9$ & $529.4$ & $488.0$ &  $529.4$ &  $233.8$ &  $485.0$ &  $286.6$ & $333.6$ & $437.7$ & $333.6$ & $\mathbf{149.7}$ &  $\mathbf{333.4}$ & $196.1$ \\
        GTSRB-CNN &  $569.0$ &  $1897.6$ & $569.0$ & $1312.7$ &  $569.0$ &  $394.0$ &  $506.0$ &  $466.4$ & $355.8$ & $1430.8$ & $355.8$ & $\mathbf{251.0}$ &  $\mathbf{355.3}$ & $330.1$ \\
        CIFAR10-FC &  $588.9$ &  $438.0$ & $588.9$ & $357.7$ & $588.9$ & $292.2$ & $582.4$ & $383.5$ & $  465.8$ & $354.9$ & $465.8$ &  $\mathbf{238.8}$& $\mathbf{465.7}$ & $316.9$ \\
        CIFAR10-CNN &  $664.8$ &  $1617.0$ & $664.8$ & $1033.5$ & $664.8$ & $448.2$ & $652.9$ & $567.8$ &  $553.5$ & $1152.8$ & $553.5$ &  $\mathbf{371.1}$ & $\mathbf{552.8}$ & $482.2$ \\ \midrule
        TaxiNet-Transformer &  $150.2$ &  $749.2$ & -- & -- & $150.2$ & $661.6$ & $150.2$ & $687.0$ & -- & -- & $91.2$ & $\mathbf{377.5}$ & $\mathbf{91.0}$ & $404.6$ \\
        IMDB-Transformer &  $20.0$ &  $161.2$ & -- & -- & $20.0$ & $30.6$ & $19.3$ & $44.3$ & -- & -- & $15.0$ & $\mathbf{37.3}$ & $\mathbf{15.0}$ & $42.3$ \\
        \bottomrule
    \end{tabular}
    }
\end{table*}

\subsection{Improvements for Explanation \emph{Size} and Generation \emph{Time}}

We report improvements in explanation \emph{size} and generation \emph{time} for image benchmarks in Table~\ref{tab:verix_verix+}, along with example explanations in Figure~\ref{fig:verix_verix+}, and statistically significant reductions in Table~\ref{tab:size-reduction}.
For each data point in the table, we collect ``valid'' explanations (i.e., excluding examples that are robust to $\magnitude$-perturbation) for the first 100 test images (to avoid selection bias) and take the average time and size. 

The overall observation is that, for the same traversal order (e.g., ``saliency''), confidence ranking reduces \emph{generation time} compared to the baseline sequential method, and the binary search-based traversal procedure further decreases the time substantially. For example, on the MNIST-CNN model, the explanation time is reduced from $439.2$ to $398.9$, and further to $46.3$, achieving an overall speedup of approximately $10\times$.
An example explanation for an MNIST image is shown in Figure~\ref{fig:verix_verix+_mnist}. For the digit ``$\mathtt{2}$'', the pixels highlighted in green are explanatory, as modifying them---particularly those in the central region---from black to white can change the model’s prediction to ``$\mathtt{8}$''.
As for \emph{explanation size}, using the same traversal procedure (e.g., ``binary search''), the bound propagation-based traversal order yields significantly smaller explanations compared to saliency-based ordering. For instance, on the GTSRB-CNN model, the average explanation size is reduced from $569.0$ to $355.8$, a reduction of approximately $38\%$.
In Figure~\ref{fig:verix_verix+_gtsrb} of the traffic sign ``$30 \ \si{mph}$'', our explanation is more focused, as it is almost entirely contained in the central region containing  ``$30$''.
Finally, in nearly all cases, QuickXplain achieves a slight additional reduction in explanation size, but at the cost of increased computation time---representing a trade-off between size and time.

\subsection{Explanations for Transformers in Real-world Applications}

We show that our approach can also be applied to \emph{transformer} models \cite{attention} and real-world applications by looking at two additional applications: a control system called TaxiNet~\cite{taxinet} for autonomous aircraft taxiing and IMDB~\cite{imdb} movie review sentiment analysis.
TaxiNet uses pictures from a camera fixed on the right wing of an aircraft to adjust steering controls to keep it on the taxiway.
Figure~\ref{fig:taxiing} shows an example explanation.
For IMDB reviews, the task is to analyze a review's sentiment ($\mathsf{positive}$ or $\mathsf{negative}$). An example explanation would be that by fixing $\mathtt{intelligent}$ and $\mathtt{movies}$ in $\mathtt{[CLS]}$ $\mathtt{one}$ $\mathtt{of}$ $\mathtt{the}$ $\mathtt{more}$ $\mathtt{intelligent}$ $\mathtt{children's}$ $\mathtt{movies}$ $\mathtt{to}$ $\mathtt{hit}$ $\mathtt{theaters}$ $\mathtt{this}$ $\mathtt{year}$, any perturbations on the embeddings of the other words will never flip the sentiment.
As IMDB are real-world reviews, their length varies, so we padded all reviews out to $1000$ tokens. In order to focus on harder problems, we report results on reviews longer than $500$ tokens (before padding).  To focus on meaningful explanations, we also exclude problems whose explanations are $50$ tokens or more.  Figure~\ref{fig:imdb} shows an example IMDB review with token-level saliency.  For both tasks, we trained transformer-based models and applied our techniques to generate formal explanations.
Table~\ref{tab:verix_verix+} shows that, also for these models, our advantages are consistent---generation time is reduced by using the binary search-inspired traversal, and explanation size is reduced by using the bound propagation-based ranking.
Note that, \emph{conf ranking} does not apply here, as the output of these two transformers is a single value. For TaxiNet, it is the cross-track distance, whereas for IMDB, it uses $\mathsf{sigmoid}$ (i.e., positive if $\geq 0.5$ and negative otherwise).

\begin{table*}[t]
\caption{Comparison between $\VeriXplus$ and other formal explainable AI approaches---\cite{izza2024distance} and \cite{huang2023robustness}---in terms of explanation size (number of pixels), generation time (seconds), and number of $\checkValid$ calls.}
\label{tab:comparison-formal}
        \centering
    \setlength{\tabcolsep}{4pt} 
    \renewcommand{\arraystretch}{1.2}
        \resizebox{\textwidth}{!}{
\begin{tabular}
{l|ccc|ccc|*{3}{>{\centering\arraybackslash}p{3.7em}}|*{3}{>{\centering\arraybackslash}p{3.7em}}}
\toprule
approaches & \multicolumn{3}{c|}{\VeriXplus} & \multicolumn{3}{c|}{\cite{izza2024distance}} & \multicolumn{3}{c|}{\cite{huang2023robustness}} & \multicolumn{3}{c}{\cite{huang2023robustness}} \\
traversal order & \multicolumn{3}{c|}{bound propagation} & \multicolumn{3}{c|}{saliency} & \multicolumn{3}{c|}{naive sequential} & \multicolumn{3}{c}{bound propagation} \\
traversal procedure & \multicolumn{3}{c|}{QuickXplain} & \multicolumn{3}{c|}{Dichotomic search} & \multicolumn{3}{c|}{QuickXplain (no optimization)} & \multicolumn{3}{c}{QuickXplain (no optimization)} \\
metrics & size & time & \# $\checkValid$ & size & time & \# $\checkValid$ & size & time & \# $\checkValid$ & size & time & \# $\checkValid$ \\
\hline \hline
MNIST-FC & $167.8$ & $35.6$ & $437.0$ & $175.9$ & $82.3$ & $1203.4$ & $265.3$ & $42.7$ & $607.7$ & $167.8$ & $45.0$ & $577.4$ \\    
MNIST-CNN & $82.4$ & $34.6$ & $211.0$ & $85.7$ & $61.3$ & $556.7$ & $158.2$ & $68.4$ & $385.8$ & $82.4$ & $82.4$ & $280.5$ \\
GTSRB-FC & $251.6$ & $140.0$ & $511.2$ & $464.3$ & $1021.1$ & $3825.6$ & $461.2$ & $190.2$ & $706.1$ & $251.6$ & $211.3$ & $761.3$ \\
GTSRB-CNN & $275.1$ & $251.0$ & $559.3$ & $479.2$ & $1672.6$ & $4040.3$ & $464.7$ & $372.4$ & $736.9$ & $275.1$ & $275.1$ & $832.7$ \\
CIFAR10-FC & $378.9$ & $251.9$ & $764.7$ & $464.7$ & $1233.4$ & $3911.5$ & $462.0$ & $227.7$ & $705.2$ & $378.9$ & $621.4$ & $1142.2$ \\
CIFAR10-CNN & $488.6$ & $478.9$ & $984.9$ & $609.3$ & $2451.6$ & $5218.9$ & $558.2$ & $430.9$ & $881.1$ & $488.6$ & $708.3$ & $1471.9$ \\
\bottomrule
\end{tabular}
}
\end{table*}

\subsection{Comparison to Existing Formal and Non-formal Explanation Approaches}

We compare $\VeriXplus$ to existing formal explainers in Table~\ref{tab:comparison-formal}.
Across all six models, we achieve \textit{smaller} explanation sizes compared to \cite{izza2024distance}, as they essentially employ saliency ranking for feature ordering. Their generation time is greater than ours, as their Dichotomic search includes \textit{unnecessary} $\checkValid$ calls, whereas our binary search-inspired traversal avoids this inefficiency. 
We also compare with \cite{huang2023robustness}. There is no indication that they use a special traversal order (perhaps because their experiments involve networks with only five features). Therefore, we first run their algorithm using a simple sequential order, and as expected, our method produces smaller explanations.  Ensuring a fair comparison, we run their algorithm using our optimized bound propagation-based traversal order, and see that our adaptation of QuickXplain is more efficient than theirs in terms of \textit{fewer} $\checkValid$ calls, and, consequently, \textit{less} generation time. We emphasize that this is not only supported by empirical evidence but also follows from a purely algorithmic analysis (see footnote~\ref{footnote:quickXplain}).

We also did a detailed comparison with non-formal explainers such as LIME~\cite{lime} and Anchors~\cite{anchors} in Tables~\ref{tab:anchors-mnist} and \ref{tab:anchors-gtsrb} of Appendix~\ref{app:lime-anchors}. We observe that our explanations are much smaller.  More significantly, we provide provable guarantees on the robustness of the produced explanations, as shown by ``robust wrt \# perturbations.'' The trade-off is that providing such guarantees requires more time.

\begin{table*}[t]
    \caption{Comparing $\VeriXplus$ explanation sizes (number of pixels) for normally (MNIST-FC) and adversarially (MNIST-FC-ADV) trained MNIST-FC models. $\magnitude$ is set to $5\%$, and $\magnitude$-robust denotes the percentage of $\magnitude$-robust samples.}
    \label{tab:adversarial}
    \centering
    \setlength{\tabcolsep}{4pt} 
    \renewcommand{\arraystretch}{1.2}
        \scalebox{0.9}{
\begin{tabular}{c|cccccccccc}
    \toprule
\multirow{3}{*}{samples} & \multicolumn{5}{c|}{MNIST-FC}                   & \multicolumn{5}{c}{MNIST-FC-ADV}       \\
                         & \multicolumn{1}{c|}{\multirow{2}{*}{accuracy}} & \multicolumn{2}{c|}{correct}           & \multicolumn{2}{c|}{incorrect}        & \multicolumn{1}{c|}{\multirow{2}{*}{accuracy}} & \multicolumn{2}{c|}{correct}           & \multicolumn{2}{c}{incorrect} \\
                         & \multicolumn{1}{c|}{}                          & $\magnitude$-robust & \multicolumn{1}{c|}{size}   & $\magnitude$-robust & \multicolumn{1}{c|}{size}   & \multicolumn{1}{c|}{}                          & $\magnitude$-robust & \multicolumn{1}{c|}{size}   & $\magnitude$-robust        & size         \\ \hline
original                   & \multicolumn{1}{c|}{$93.76\%$}                   & $4\%$     & \multicolumn{1}{c|}{$177.20$} & $0.3\%$   & \multicolumn{1}{c|}{$398.85$} & \multicolumn{1}{c|}{$92.85\%$}                   & $52.3\%$  & \multicolumn{1}{c|}{$128.22$} & $2\%$            & $311.14$       \\
malicious       & \multicolumn{1}{c|}{$23.66\%$}                   & $0\%$     & \multicolumn{1}{c|}{$466.30$} & $0\%$     & \multicolumn{1}{c|}{$562.07$} & \multicolumn{1}{c|}{$82.66\%$}                   & $6.7\%$   & \multicolumn{1}{c|}{$298.57$} & $0.3\%$          & $536.41$       \\
    \bottomrule
\end{tabular}
}
\end{table*}

\subsection{Explanations from Adversarial Training}
\label{sec:adversarial}

In Table~\ref{tab:adversarial}, we compare our explanations for normally and \emph{adversarially} trained MNIST-FC models on original and \emph{malicious} samples. 
We describe the adversarial training process in Appendix~\ref{app:adv_training}.
We produce explanations for both \emph{correct} and \emph{incorrect} samples. For each table entry, we collected $300$ samples and report their average explanation size (excluding $\magnitude$-robust samples).
Overall, we observe that both models have smaller explanations for original samples than for malicious samples, and for correct predictions than for incorrect predictions. Notably, the MNIST-FC-ADV model produces smaller explanations than MNIST-FC under all conditions. For instance, for original, correct samples, MNIST-FC-ADV produces $27.64\%$ smaller explanations ($128.22$~vs.~$177.20$). This suggests that with adversarial training, the model learns more implicit information about the samples, e.g., which pixels likely contain the key information for prediction and which are subject to trivial perturbations; thus, it only needs to pay attention to fewer pixels to make a correct decision. In Figure~\ref{fig:adv-examples} of Appendix~\ref{app:adv_training}, we show examples of both original and malicious inputs. 
A similar phenomenon is observable in the $\magnitude$-robust rate, which increases from $4\%$ to $52.3\%$ after adversarial training, since now the model has learned to focus on the principal features in the input and to become less sensitive to perturbations.

\begin{figure}[t]
    \centering
    \begin{subfigure}{\linewidth}
        \centering
        \includegraphics[height=3.4cm]{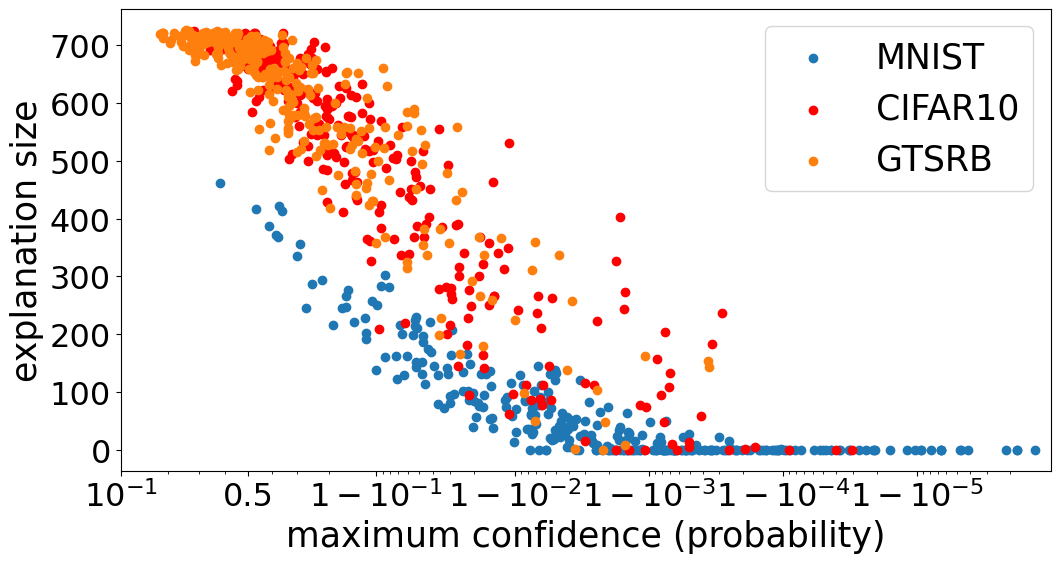}
        \caption{MNIST, CIFAR10, GTSRB: confidence \& size}
        \label{fig:ood-cnn-prob-size}
    \end{subfigure}
    \begin{subfigure}{0.48\linewidth}
        \centering
        \includegraphics[height=3.47cm]{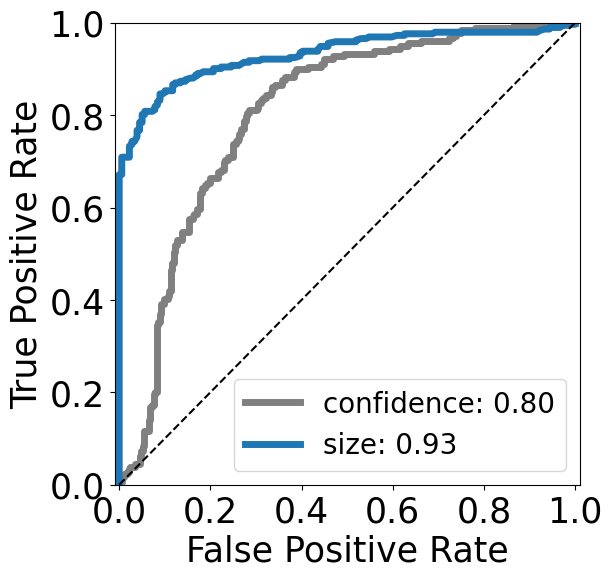}
        \caption{CIFAR10: AUROC}
        \label{fig:ood-cnn-cifar10-auroc}
    \end{subfigure}
    \begin{subfigure}{0.48\linewidth}
        \centering
        \includegraphics[height=3.47cm]{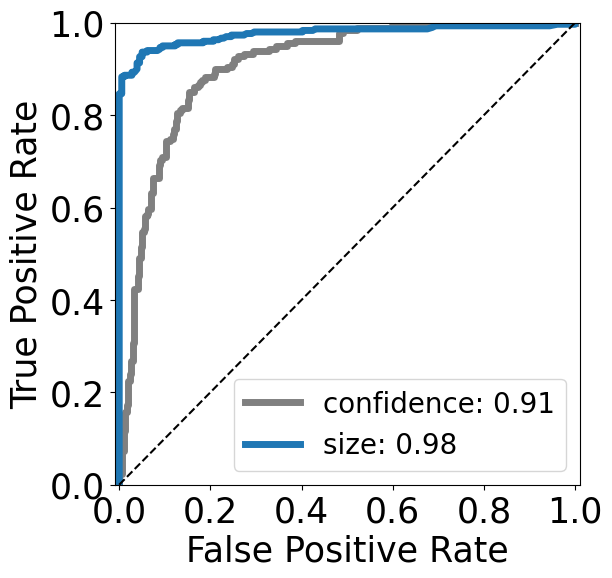}
        \caption{GTSRB:AUROC}
        \label{fig:ood-cnn-gtsrb-auroc}
    \end{subfigure}
    \caption{Detecting \emph{out-of-distribution} examples from CIFAR10 and GTSRB for the MNIST-CNN model. (a) Explanation size and maximum confidence (\emph{log} scale, see \emph{linear} scale in Figure~\ref{fig:ood_cnn_linear}). (b)(c) ROC curves and AUROC values for OOD samples from CIFAR10 and GTSRB, respectively.}
    \label{fig:ood-cnn}
\end{figure}

\subsection{Using Explanation Size to Detect OOD and Misclassified Examples}
\label{sec:ood}

We also show that explanation size can help detect \emph{out-of-distribution} (OOD) samples. Consider a scenario in which CIFAR10 and GTSRB images are fed into the MNIST-CNN model as OOD samples. We crop the images so they have the same size as MNIST images and use the $\mathsf{OpenCV}$~\cite{opencv_library}\footnote{\url{https://pypi.org/project/opencv-python/}} library to convert color images to grayscale. The goal is to preserve the primary semantic meanings at the center of these images.
We collected $900$ samples in total---$300$ each from MNIST, CIFAR10, and GTSRB, and plot their maximum softmax probability and explanation size, as shown in Figure~\ref{fig:ood-cnn-prob-size} (see also Appendix~\ref{app:ood}). We observe a significant separation between the in- and out-of-distribution samples, suggesting that smaller explanation sizes are associated with in-distribution inputs. A standard technique from previous work~\cite{softmax-baseline} uses the maximum softmax probabilities for OOD detection. 
We compare using explanation size and confidence to detect OOD samples from CIFAR10 and GTSRB. From Figures~\ref{fig:ood-cnn-cifar10-auroc} and \ref{fig:ood-cnn-gtsrb-auroc}, we see that explanation size yields better ROC curves for both datasets and also achieves higher AUROC values, i.e., $93\%$ on CIFAR10 and $98\%$ on GTSRB. We perform similar OOD detection on the MNIST-FC model in
Appendix~\ref{app:ood}. We remark that this section aims to illustrate the utility of formal explanations, not to outperform state-of-the-art OOD detection methods, which are beyond this paper's scope.
In Appendix~\ref{app:incorrect}, we also show that explanation size is a useful proxy for detecting \emph{misclassified} samples.

\section{Related Work}
\label{sec:related}

Several approaches to formal explanations~\cite{formalXAI} have been explored recently.
\cite{abduction} first proposed using abductive reasoning to compute formal explanations for neural networks by encoding them into a set of constraints and then deploying automated reasoning systems such as SMT solvers to solve the constraints. 
\cite{ore} brings in bounded perturbations, $k$-nearest neighbors and $\magnitude$-balls to produce distance-restricted explanations for natural language models. 
Adapting the $\magnitude$-perturbations to perception models for which inputs tend to be naturally bounded, \cite{verix} proposes a feature-level saliency to obtain a ranking of the input features and thus produce empirically small explanations.
In this paper, we utilize bound propagation-based techniques to obtain more fine-grained feature-level sensitivity. Compared to the existing saliency from \cite{verix} and later adopted by \cite{izza2024distance}, we obtain $\magnitude$-perturbation-dependent traversal orders that lead to even smaller explanation sizes.
To reduce generation time, \cite{huang2023robustness} mimics the QuickXplain algorithm~\cite{quickxplain} to avoid traversing all the features in a linear way (experiments only include the linear case though). 
Our optimization of this QuickXplain adaptation further reduces the number of oracle calls; we also perform a complete experimental comparison between the linear and non-linear approaches.
Additionally, we introduce binary search-based traversals to further improve time. The Dichotomic search method from \cite{izza2024distance} is also inspired by binary search, but it includes unnecessary oracle calls, as it restarts the search from the beginning every time it finds a ``transition feature.'' Therefore, the number of oracle calls needed by their Dichotomic search in their experiments is ``always larger than'' the number of input features. In contrast, our binary search-based algorithm does not have such redundant oracle calls and thus achieves much reduced time.
Finally, our confidence ranking strategy accelerates the $\checkValid$ procedure, which benefits all such oracle calls.
To the best of our knowledge, this is the \textit{first} time that this strategy has been proposed.


\section{Discussion of Limitations}

While our approach achieves improved scalability compared to existing formal explainers, we note that—with strict soundness and completeness guarantees—scalability is still limited.  In particular, our method does not yet scale to contemporary large models, including language models. We emphasize that this limitation arises from the inherent complexity of the underlying problem.  Finding ways to help address this gap remains an important direction for future work.
Moreover, although we use explanation size (e.g., number of pixels for images or tokens for texts) as a proxy for quality, there is no consensus on whether this is the most appropriate metric. Exploring alternative evaluation methods, such as human-in-the-loop assessments, would be a promising avenue for further investigation.

\section{Conclusion and Future Work}
\label{sec:conclusion}

We have presented the $\VeriXplus$ framework for computing optimal verified explanations with improved size and generation time. 
%
Future work could explore further techniques for improving the performance of verified explanation generation, perhaps by adapting parallel techniques from~\cite{izza2024distance} or by finding ways to introduce approximations in order to gain scalability.  It would also be interesting to evaluate explanations using a controlled setting \cite{tcav} rather than simply size.  
Finally, we would like to explore additional applications, especially in areas where safety or fairness is crucial.

\section*{Acknowledgments}

This work was funded in part by IBM as a founding member of the Stanford Institute for Human-centered Artificial Intelligence (HAI), a Ford Alliance Project (316280), the National Science Foundation (grant number 2211505), the Stanford CURIS program, and the Stanford Center for AI Safety.


\bibliography{aaai2026}

@article{opencv_library,
  author = {Bradski, Gary},
  title = {The OpenCV Library},
  journal = {Dr. Dobb's Journal of Software Tools},
  year = {2000}
}

@inproceedings{verix,
 author = {Wu, Min and Wu, Haoze and Barrett, Clark},
 booktitle = {Advances in Neural Information Processing Systems},
 editor = {A. Oh and T. Naumann and A. Globerson and K. Saenko and M. Hardt and S. Levine},
 pages = {22247--22268},
 publisher = {Curran Associates, Inc.},
 title = {{VeriX}: Towards Verified Explainability of Deep Neural Networks},
 url = {https://proceedings.neurips.cc/paper_files/paper/2023/file/46907c2ff9fafd618095161d76461842-Paper-Conference.pdf},
 volume = {36},
 year = {2023}
}

@inproceedings{formalXAI, 
title={Delivering Trustworthy AI through Formal XAI}, 
volume={36},
url={https://ojs.aaai.org/index.php/AAAI/article/view/21499},
DOI={10.1609/aaai.v36i11.21499}, 
number={11}, 
booktitle={Proceedings of the AAAI Conference on Artificial Intelligence},
author={Marques-Silva, Joao and Ignatiev, Alexey}, 
year={2022}, 
month={Jun.},
pages={12342-12350}
}

@inproceedings{quickxplain,
  title={Quickxplain: Preferred explanations and relaxations for over-constrained problems},
  author={Junker, Ulrich},
  booktitle={Proceedings of the 19th national conference on Artifical intelligence},
  pages={167--172},
  year={2004}
}

@inproceedings{abduction,
  title={Abduction-based explanations for machine learning models},
  author={Ignatiev, Alexey and Narodytska, Nina and Marques-Silva, Joao},
  booktitle={Proceedings of the AAAI Conference on Artificial Intelligence},
  pages={1511--1519},
  year={2019}
}

@inproceedings{PI-explanations,
  title={A symbolic approach to explaining Bayesian network classifiers},
  author={Shih, Andy and Choi, Arthur and Darwiche, Adnan},
  booktitle={Proceedings of the 27th International Joint Conference on Artificial Intelligence},
  pages={5103--5111},
  year={2018}
}

@inproceedings{sufficient-reasons,
  title={On the reasons behind decisions},
  author={Darwiche, Adnan and Hirth, Auguste},
  booktitle={Proceedings of the 24th European Conference on Artificial Intelligence},
  year={2020}
}

@inproceedings{ore,
  title     = {On Guaranteed Optimal Robust Explanations for NLP Models},
  author    = {La Malfa, Emanuele and Michelmore, Rhiannon and Zbrzezny, Agnieszka M. and Paoletti, Nicola and Kwiatkowska, Marta},
  booktitle = {Proceedings of the Thirtieth International Joint Conference on
               Artificial Intelligence, {IJCAI-21}},
  editor    = {Zhi-Hua Zhou},
  pages     = {2658--2665},
  year      = {2021},
  month     = {8},
  doi       = {10.24963/ijcai.2021/366},
}

@inproceedings{izza2024distance,
  author       = {Yacine Izza and Xuanxiang Huang and António Morgado and Jordi Planes and Alexey Ignatiev and Jo{\~a}o Marques-Silva},
  title        = {Distance-Restricted Explanations: Theoretical Underpinnings \& Efficient Implementation},
  booktitle    = {Proceedings of the 21st International Conference on Principles of Knowledge Representation and Reasoning (KR 2024)},
  pages        = {475--486},
  year         = {2024},
  publisher    = {Association for the Advancement of Artificial Intelligence},
  doi          = {10.24963/kr.2024/45},
  url          = {https://arxiv.org/abs/2405.08297}
}

@article{huang2023robustness,
  title={From robustness to explainability and back again},
  author={Huang, Xuanxiang and Marques-Silva, Joao},
  journal={arXiv preprint arXiv:2306.03048},
  year={2023}
}

@inproceedings{anchors,
  title={Anchors: high-precision model-agnostic explanations},
  author={Ribeiro, Marco Tulio and Singh, Sameer and Guestrin, Carlos},
  booktitle={Proceedings of the Thirty-Second AAAI Conference on Artificial Intelligence and Thirtieth Innovative Applications of Artificial Intelligence Conference and Eighth AAAI Symposium on Educational Advances in Artificial Intelligence},
  pages={1527--1535},
  year={2018}
}

@inproceedings{lime,
  title={``Why should i trust you?'' Explaining the predictions of any classifier},
  author={Ribeiro, Marco Tulio and Singh, Sameer and Guestrin, Carlos},
  booktitle={Proceedings of the 22nd ACM SIGKDD international conference on knowledge discovery and data mining},
  pages={1135--1144},
  year={2016}
}

@inproceedings{shap,
  title={A unified approach to interpreting model predictions},
  author={Lundberg, Scott M and Lee, Su-In},
  booktitle={Proceedings of the 31st International Conference on Neural Information Processing Systems},
  pages={4768--4777},
  year={2017}
}

@inproceedings{ibp,
  title={Scalable verified training for provably robust image classification},
  author={Gowal, Sven and Dvijotham, Krishnamurthy Dj and Stanforth, Robert and Bunel, Rudy and Qin, Chongli and Uesato, Jonathan and Arandjelovic, Relja and Mann, Timothy and Kohli, Pushmeet},
  booktitle={Proceedings of the IEEE/CVF International Conference on Computer Vision},
  pages={4842--4851},
  year={2019}
}

@misc{Mathiesen2022,
  author = {Frederik Baymler Mathiesen},
  title = {Bound Propagation},
  year = {2022},
  publisher = {GitHub},
  journal = {GitHub repository},
  howpublished = {\url{https://github.com/Zinoex/bound_propagation}}
}

@inproceedings{crown,
  title={Efficient neural network robustness certification with general activation functions},
  author={Zhang, Huan and Weng, Tsui-Wei and Chen, Pin-Yu and Hsieh, Cho-Jui and Daniel, Luca},
  booktitle={Proceedings of the 32nd International Conference on Neural Information Processing Systems},
  pages={4944--4953},
  year={2018}
}

@inproceedings{bcrown,
  title={Beta-CROWN: Efficient Bound Propagation with Per-neuron Split Constraints for Neural Network Robustness Verification},
  author={Wang, Shiqi and Zhang, Huan and Xu, Kaidi and Lin, Xue and Jana, Suman and Hsieh, Cho-Jui and Kolter, Zico},
  booktitle={Proceedings of the 35th Conference on Neural Information Processing Systems (NeurIPS)},
  volume={34},
  pages={29909--29921},
  year={2021}
}

@inproceedings{marabou,
  title={The marabou framework for verification and analysis of deep neural networks},
  author={Katz, Guy and Huang, Derek A and Ibeling, Duligur and Julian, Kyle and Lazarus, Christopher and Lim, Rachel and Shah, Parth and Thakoor, Shantanu and Wu, Haoze and Zelji{\'c}, Aleksandar and others},
  booktitle={International Conference on Computer Aided Verification},
  pages={443--452},
  year={2019},
}

@article{prima,
  title={PRIMA: general and precise neural network certification via scalable convex hull approximations},
  author={M{\"u}ller, Mark Niklas and Makarchuk, Gleb and Singh, Gagandeep and P{\"u}schel, Markus and Vechev, Martin},
  journal={Proceedings of the ACM on Programming Languages},
  volume={6},
  number={POPL},
  pages={1--33},
  year={2022},
  publisher={ACM New York, NY, USA}
}

@article{deeppoly,
  title={An abstract domain for certifying neural networks},
  author={Singh, Gagandeep and Gehr, Timon and P{\"u}schel, Markus and Vechev, Martin},
  journal={Proceedings of the ACM on Programming Languages},
  volume={3},
  number={POPL},
  pages={1--30},
  year={2019},
  publisher={ACM New York, NY, USA}
}

@inproceedings{verinet,
  author    = {Patrick Henriksen and Alessio R. Lomuscio},
  title     = {Efficient Neural Network Verification via Adaptive Refinement and Adversarial Search},
  booktitle = {Proceedings of the 24th European Conference on Artificial Intelligence (ECAI 2020)},
  pages     = {2513--2520},
  year      = {2020},
  doi       = {10.3233/FAIA200385}
}

@inproceedings{marabou2,
  title={Marabou 2.0: a versatile formal analyzer of neural networks},
  author={Wu, Haoze and Isac, Omri and Zelji{\'c}, Aleksandar and Tagomori, Teruhiro and Daggitt, Matthew and Kokke, Wen and Refaeli, Idan and Amir, Guy and Julian, Kyle and Bassan, Shahaf and others},
  booktitle={International Conference on Computer Aided Verification},
  pages={249--264},
  year={2024},
  organization={Springer}
}

@article{mnist,
  title={MNIST handwritten digit database},
  author={LeCun, Yann and Cortes, Corinna and Burges, CJ},
  journal={ATT Labs [Online]. Available: http://yann.lecun.com/exdb/mnist},
  volume={2},
  year={2010}
}

@article{gtsrb,
  title={Man vs. computer: Benchmarking machine learning algorithms for traffic sign recognition},
  author={Stallkamp, Johannes and Schlipsing, Marc and Salmen, Jan and Igel, Christian},
  journal={Neural networks},
  volume={32},
  pages={323--332},
  year={2012},
  publisher={Elsevier}
}

@TECHREPORT{cifar10,
  title={Learning multiple layers of features from tiny images},
  author={Krizhevsky, Alex and others},
  year={2009}
}

@inproceedings{imdb,
    title = "Learning Word Vectors for Sentiment Analysis",
    author = "Maas, Andrew L.  and
      Daly, Raymond E.  and
      Pham, Peter T.  and
      Huang, Dan  and
      Ng, Andrew Y.  and
      Potts, Christopher",
    editor = "Lin, Dekang  and
      Matsumoto, Yuji  and
      Mihalcea, Rada",
    booktitle = "Proceedings of the 49th Annual Meeting of the Association for Computational Linguistics: Human Language Technologies",
    month = jun,
    year = "2011",
    address = "Portland, Oregon, USA",
    publisher = "Association for Computational Linguistics",
    url = "https://aclanthology.org/P11-1015/",
    pages = "142--150"
}

@inproceedings{bound_propagation,
author = {Xu, Kaidi and Shi, Zhouxing and Zhang, Huan and Wang, Yihan and Chang, Kai-Wei and Huang, Minlie and Kailkhura, Bhavya and Lin, Xue and Hsieh, Cho-Jui},
title = {Automatic perturbation analysis for scalable certified robustness and beyond},
year = {2020},
isbn = {9781713829546},
publisher = {Curran Associates Inc.},
address = {Red Hook, NY, USA},
booktitle = {Proceedings of the 34th International Conference on Neural Information Processing Systems},
articleno = {96},
numpages = {13},
location = {Vancouver, BC, Canada},
series = {NIPS '20}
}

@inproceedings{
softmax-baseline,
title={A Baseline for Detecting Misclassified and Out-of-Distribution Examples in Neural Networks},
author={Dan Hendrycks and Kevin Gimpel},
booktitle={International Conference on Learning Representations},
year={2017},
url={https://openreview.net/forum?id=Hkg4TI9xl}
}

@INPROCEEDINGS{AI2,
  author={Gehr, Timon and Mirman, Matthew and Drachsler-Cohen, Dana and Tsankov, Petar and Chaudhuri, Swarat and Vechev, Martin},
  booktitle={2018 IEEE Symposium on Security and Privacy (SP)}, 
  title={AI2: Safety and Robustness Certification of Neural Networks with Abstract Interpretation}, 
  year={2018},
  pages={3-18},
  doi={10.1109/SP.2018.00058}}

@inproceedings{reluplex,
  title={Reluplex: An efficient SMT solver for verifying deep neural networks},
  author={Katz, Guy and Barrett, Clark and Dill, David L and Julian, Kyle and Kochenderfer, Mykel J},
  booktitle={International conference on computer aided verification},
  pages={97--117},
  year={2017},
  organization={Springer}
}

@inproceedings{ood-confidence,
  title={Why relu networks yield high-confidence predictions far away from the training data and how to mitigate the problem},
  author={Hein, Matthias and Andriushchenko, Maksym and Bitterwolf, Julian},
  booktitle={Proceedings of the IEEE/CVF conference on computer vision and pattern recognition},
  pages={41--50},
  year={2019}
}

@article{goodfellow2014explaining,
  title={Explaining and harnessing adversarial examples},
  author={Goodfellow, Ian J and Shlens, Jonathon and Szegedy, Christian},
  journal={arXiv preprint arXiv:1412.6572},
  year={2014}
}

@inproceedings{guo2017calibration,
author = {Guo, Chuan and Pleiss, Geoff and Sun, Yu and Weinberger, Kilian Q.},
title = {On calibration of modern neural networks},
year = {2017},
publisher = {JMLR.org},
booktitle = {Proceedings of the 34th International Conference on Machine Learning - Volume 70},
pages = {1321–1330},
numpages = {10},
location = {Sydney, NSW, Australia},
series = {ICML'17}
}

@inproceedings{taxinet,
  title={Validation of image-based neural network controllers through adaptive stress testing},
  author={Julian, Kyle D and Lee, Ritchie and Kochenderfer, Mykel J},
  booktitle={2020 IEEE 23rd international conference on intelligent transportation systems (ITSC)},
  pages={1--7},
  year={2020},
  organization={IEEE}
}

@inproceedings{attention,
author = {Vaswani, Ashish and Shazeer, Noam and Parmar, Niki and Uszkoreit, Jakob and Jones, Llion and Gomez, Aidan N. and Kaiser, \L{}ukasz and Polosukhin, Illia},
title = {Attention is all you need},
year = {2017},
isbn = {9781510860964},
publisher = {Curran Associates Inc.},
address = {Red Hook, NY, USA},
booktitle = {Proceedings of the 31st International Conference on Neural Information Processing Systems},
pages = {6000–6010},
numpages = {11},
location = {Long Beach, California, USA},
series = {NIPS'17}
}

@inproceedings{tcav,
  title={Interpretability beyond feature attribution: Quantitative testing with concept activation vectors (tcav)},
  author={Kim, Been and Wattenberg, Martin and Gilmer, Justin and Cai, Carrie and Wexler, James and Viegas, Fernanda and others},
  booktitle={International conference on machine learning},
  pages={2668--2677},
  year={2018},
  organization={PMLR}
}

\newpage

\appendix
\newpage
~\newpage

\section{Proofs for Theorems in Section~\ref{sec:methodology}}
\label{app:proofs}

\subsection{Proof for Theorem~\ref{thm:timeBinary}}
\label{app:proofs-1}

\begin{reptheorem}{thm:timeBinary}[Time Complexity]

    Given a neural network $\network$ and an input $\image = \langle \pixel_1, \ldots, \pixel_\inDim \rangle$ where $\inDim \geq 2$, the \emph{time complexity} of $\binaryCompute(\network, \image)$ is $2$ calls of $\checkValid$ for the \emph{best} case (all features are irrelevant) and $k_{2\inDim} = 2 \cdot k_\inDim + 1$ or $k_{2\inDim+1} = k_{\inDim+1} + k_\inDim + 1$, for which $k_2=2$ and $k_3=4$ are the base cases, calls of $\checkValid$ for the \emph{worst} case (all features are explanatory). When $\inDim = 1$, it is obvious that only 1 $\checkValid$ call is needed. 
\end{reptheorem}

\begin{proof}
    It is straightforward that, when $\image$ has a single feature, i.e., $\inDim = 1$ and thus $\abs{\image_\indexSet}=1$, only the first $\mathbf{if}$ condition of Algorithm~\ref{alg:binaryComputation} will be executed. That said, only $1$ call of the $\checkValid$ procedure in Line~\ref{line:singleCheck} is needed.

    For the non-base case, when there are more than one input features, i.e., $\inDim \geq 2$, we analyze the time complexity for the \emph{best} case and the \emph{worst} case separately.

    \begin{itemize}[topsep=3pt,itemsep=4pt,leftmargin=10pt]
        
    \item In the \emph{best} case, all the features are irrelevant so the input is essentially $\magnitude$-robust. The original input $\image$ is split into two subsets $\image_\Phi$ and $\image_\Psi$ in Line~\ref{line:split}. Algorithm~\ref{alg:binaryComputation} then terminates after two calls of $\checkValid$ in Lines~\ref{line:phiCheck} and \ref{line:psiCheck} to examine whether $\image_\Phi$ and $\image_\Psi$ are irrelevant, respectively. When $\checkValid$ returns $\true$, they are put into the irrelevant set $\image_\irrelevant$ and the algorithm terminates. Therefore, for the best case, only $2$ $\checkValid$ calls are needed.
    
    \item In the \emph{worst} case, all the input features are relevant. That is, the whole input $\image$ is an explanation. 
    For this, we use $k_\inDim$ to denote the number of $\checkValid$ calls needed for an input that comprises $\inDim$ features, and use $k_{2\inDim}$ and $k_{2\inDim+1}$ to denote that of an input comprising even and odd features, respectively. 
    
    For an input $\image$ that has $2\inDim$ features (i.e., \emph{even} features), when calling $\binaryCompute$, Line~\ref{line:split} splits it into two subsets $\image_\Phi$ and $\image_\Psi$, each of which has $\inDim$ features. Then, the algorithm runs the $\checkValid$ procedure in  Line~\ref{line:phiCheck} to check if the first subset $\image_\Phi$ is irrelevant. It returns $\false$ as all features are relevant. Thus, the algorithm proceeds to the $\mathbf{else}$ condition in Line~\ref{line:phiFalse}. Since $\image_\Phi$ has $\inDim$ features and $\inDim \geq 2$, $\binaryCompute(\network, \image_\Phi)$ in Line~\ref{line:phiBinary} is invoked. Afterwards, $\binaryCompute(\network, \image_\Psi)$ is invoked to check the second subset $\image_\Psi$. Each such instantiation of $\binaryCompute$ takes the worst case as the whole input is an explanation. Therefore, for an input comprising $2\inDim$ features, it will need $k_\inDim$ $\checkValid$ calls for subset $\image_\Phi$ and $k_\inDim$ $\checkValid$ calls for subset $\image_\Psi$, as well as the extra $1$ $\checkValid$ call in Line~\ref{line:phiFalse} at the beginning. That is, $k_{2\inDim} = 2 \cdot k_\inDim + 1$. This is the case when the input has even number of features. 
    
    Now if it has \emph{odd} features, e.g., $2\inDim + 1$, it will be split into two subsets with different sizes. In our case, when $\abs{\image}$ is odd, we set $\abs{\image_\Phi} = \abs{\image_\Psi} + 1$. That is, the first subset $\image_\Phi$ has $\inDim + 1$ features and the second subset $\image_\Psi$ has $\inDim$ features. Similarly, the algorithm needs to run the $1$ $\checkValid$ in Line~\ref{line:phiCheck}, and then proceeds to work with $\image_\Phi$ and $\image_\Psi$ accordingly. Therefore, for an input that consists of $2\inDim + 1$ features, we have $k_{2\inDim + 1} = k_{\inDim + 1} + k_\inDim + 1$.

    \end{itemize}
     
\end{proof}

\newpage

\subsection{Proof for Theorem~\ref{thm:timeQuickXplain}}
\label{app:proofs-2}

\begin{reptheorem}{thm:timeQuickXplain}[Time Complexity]
    Given a neural network $\network$ and an input $\image = \langle \pixel_1, \ldots, \pixel_\inDim \rangle$ where $\inDim \geq 2$, the \emph{time complexity} of $\qXp(\emptyset, \emptyset, \image)$ is $\lfloor \log_2 \inDim \rfloor + 1$ calls of $\checkValid$ in the \emph{best} case (all features are irrelevant) and $(\inDim-1) \times 2$ calls of $\checkValid$ in the \emph{worst} case (all features are explanatory). When $\inDim = 1$, it is obvious that only 1 $\checkValid$ call is needed. 
\end{reptheorem}

\begin{proof}

    It is straightforward that, when $\image$ has a single feature, i.e., $\inDim = 1$ and thus $\abs{\image_\indexSet}=1$, only the first $\mathbf{if}$ condition of Algorithm~\ref{alg:QuickXplain} will be executed. That said, only $1$ call of the $\checkValid$ procedure in Line~\ref{line:QXPsingleX} is needed.

    For the non-base case, when there are more than one input features, i.e., $\inDim \geq 2$, we analyze the time complexity of $\qXp(\emptyset, \emptyset, \image)$ for the \emph{best} case and the \emph{worst} case separately.

    \begin{itemize}[topsep=3pt,itemsep=4pt,leftmargin=10pt]

    \item In the \emph{best} case, all features are irrelevant so the input is essentially $\magnitude$-robust. The original input $\image$ is split into two subsets $\image_\Phi$ and $\image_\Psi$ in Line~\ref{line:QXPsplit}. Then in Line~\ref{line:QXPphiIf} the $\checkValid$ procedure examines if the first subset $\image_\Phi$ is irrelevant. Since this is the best case when all features are irrelevant, it will return $\true$. Then Line~\ref{line:QXPpsi} will be executed to recursively call the $\qXp$ function to process the second subset $\image_\Psi$. In the new instantiation, similarly Lines~\ref{line:QXPsplit}, \ref{line:QXPphiIf}, and \ref{line:QXPpsi} will be executed so the $\qXp$ function is revoked again. This recursion continues until there is only one feature in the candidate feature set $\image_\indexSet$. Then in Line~\ref{line:QXPCheckSingle}, the $\checkValid$ procedure examines the single feature, and the algorithm returns. That said, the number of $\checkValid$ calls is how many times the original input $\image$ can be divided into two subsets until there is only one feature left, plus the final $\checkValid$ call to process the last single feature. In other words, it is the number of times we can divide $\inDim$ by $2$ until we get $1$, which is $\lfloor \log_2 \inDim \rfloor$, plus the extra $1$ call, hence  $\lfloor \log_2 \inDim \rfloor + 1$. Here, the floor function $\lfloor \cdot \rfloor$ is to accommodate the situation when $\log_2 \inDim$ is not an integer, since we split $\image$ as $\abs{\image_\Phi} = \abs{\image_\Psi} + 1$ when $\image$ has odd features.


    \item In the \emph{worst} case, all the input features are relevant. That is, the whole input $\image$ is an explanation. 
    After splitting $\image$ into two subsets $\image_\Phi$ and $\image_\Psi$ in Line~\ref{line:QXPsplit}, the $\checkValid$ procedure examines whether $\image_\Phi$ and $\image_\Psi$ are irrelevant features in the $\mathbf{if}$ conditions in Lines~\ref{line:QXPphiIf} and \ref{line:QXPpsiIf}, respectively. Both will return $\false$ as all input features are relevant for this case. Then the $\qXp$ function is recursively called to process the first subset $\image_\Phi$ in Line~\ref{line:newQXPphi} and the second subset $\image_\Psi$ in Line~\ref{line:newQXPpsi}. These new instantiations will have similar executions as above, until eventually each subset contains only a single feature. Then, in Lines~\ref{line:QXP1phi} and \ref{line:QXP1psi}, subsets $\image_\Phi$ and $\image_\Psi$ will be directly regarded as explanation features. Note that here we do not need another round of calling the $\qXp$ function, as we already know from Lines~\ref{line:QXPphiIf} and \ref{line:QXPpsiIf} that $\image_\Phi$ and $\image_\Psi$ are not irrelevant thus in the explanation. That said, every time the candidate set $\image_\indexSet$ is split into two subsets, two $\checkValid$ calls are needed for both subsets. Such division terminates when each input feature is in an individual subset. Since for an input $\image$ that has $\inDim$ features, $\inDim - 1$ divisions are needed, therefore the total number of $\checkValid$ calls is $(\inDim - 1) \cdot 2$. 
    
    
\end{itemize}

\end{proof}

\newpage
\clearpage

\section{Supplementary Experimental Results}
\label{app:experiments}

Due to space limitations, we provide additional experimental results that supplement Section~\ref{sec:experiments}.

\subsection{Improvements for Explanation Size and Generation Time}
\label{app:size-time}

Figure~\ref{fig:verix_verix+} presents example explanations generated by $\VeriX$ and $\VeriXplus$ on the MNIST, GTSRB, and CIFAR10 datasets. Among them, the GTSRB examples show the most noticeable improvement. Table~\ref{tab:size-reduction} provides supporting experimental evidence: specifically, the “size reduction” column shows that $\VeriXplus$ achieves the highest reduction—approximately $37\%$—on both the fully connected and convolutional GTSRB models. This aligns with the stronger visual improvements observed for GTSRB compared to MNIST and CIFAR10.

\subsection{Explanations for Transformers and Real-world Applications}

In the bottom two rows of Table~\ref{tab:verix_verix+}, we demonstrate that our $\VeriXplus$ framework extends to transformers and real-world applications, including autonomous aircraft taxiing (TaxiNet-Transformer) and sentiment analysis (IMDB-Transformer). While TaxiNet and other standard benchmarks in Table~\ref{tab:verix_verix+} involve image inputs, the IMDB dataset consists of natural language movie reviews.

We now provide some intuition on how our approach applies to natural language processing (NLP) models. In this context, a text is treated analogously to an image. After tokenization, each token is conceptually similar to a pixel: both are represented as real-valued vectors—token embeddings for text and RGB values for color images. Given a text input and a transformer-based NLP model, we apply either the saliency method or our bound propagation-based method to obtain a token traversal order. We then use one of our traversal procedures—sequential, binary search, or QuickXplain—to partition the tokens into two disjoint sets: explanatory and irrelevant tokens. The explanation size is defined as the number of explanatory tokens.

\begin{figure*}[t]
    \centering
    \includegraphics[angle=90,width=0.65\linewidth]{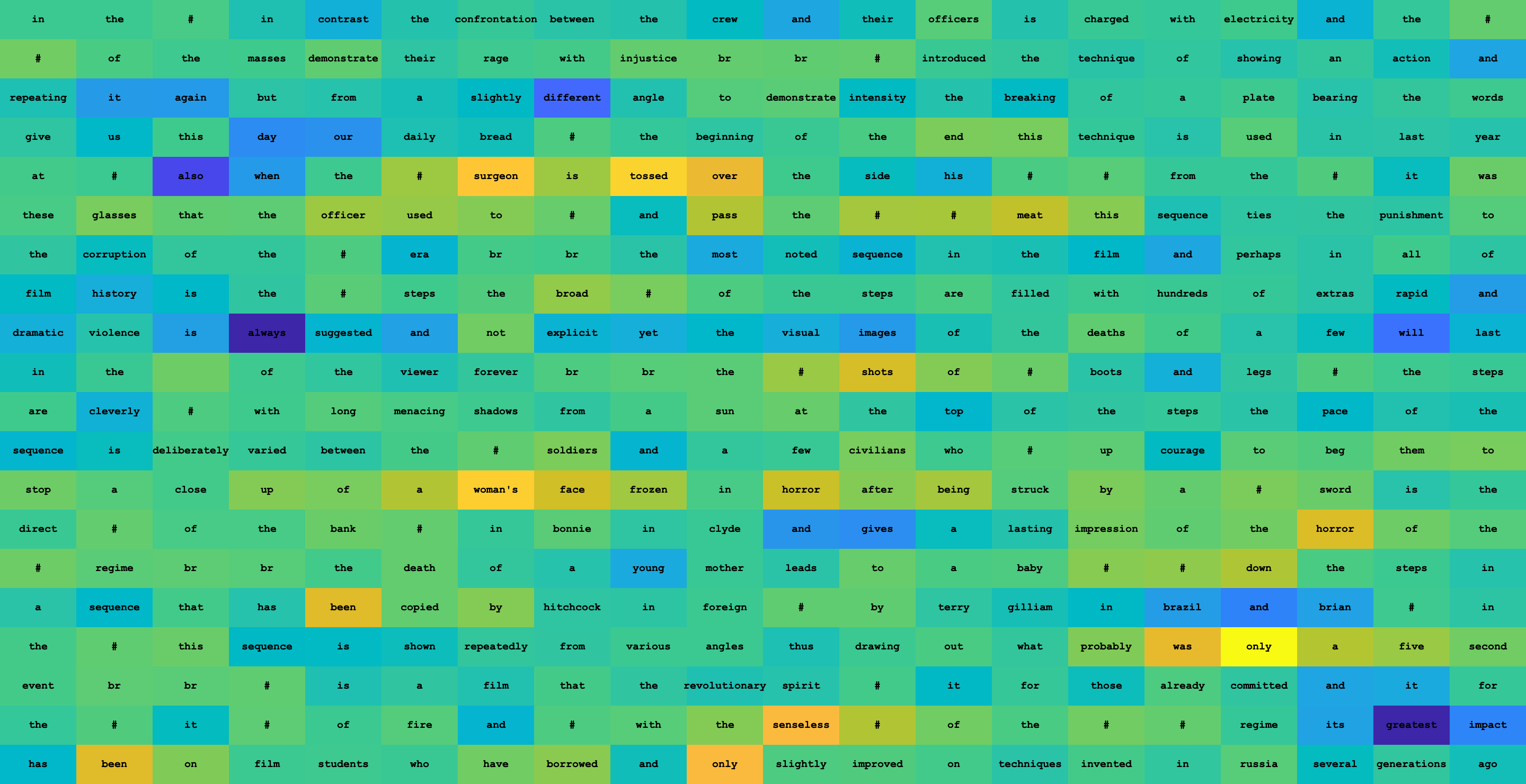}
    \caption{An example review from the IMDB dataset with token-level sensitivity computed from the saliency method. For instance, tokens such as ``$\mathtt{always}$'' and ``$\mathtt{greatest}$'' (highlighted in blue) bring in a positive tone while tokens such as ``$\mathtt{surgeon}$'' and ``$\mathtt{horror}$'' (highlighted in yellow) imply a negative tone.}
    \label{fig:imdb}
\end{figure*}

Figure~\ref{fig:imdb} shows an example IMDB review with token-level sensitivity scores computed using the saliency method. For instance, positive words such as “$\mathtt{always}$” and “$\mathtt{greatest}$” are highlighted in blue, while negative words such as “$\mathtt{surgeon}$” and “$\mathtt{horror}$” appear in yellow. By ranking tokens based on their sensitivity scores, we obtain a traversal order, which is then used to compute the set of explanatory tokens with respect to the given text and transformer model.

\subsection{Comparison to Existing Formal Explanation Approaches}
\label{app:comparison-formal}

In this section, we compare our framework, $\VeriXplus$, with existing formal explanation methods~\cite{izza2024distance,huang2023robustness,ore,abduction}. Specifically, Table~\ref{tab:comparison-formal} presents a comparison with the methods of \cite{izza2024distance} and \cite{huang2023robustness}, using three metrics: explanation size, generation time, and the number of oracle calls, i.e., $\# \checkValid$.

Compared to \cite{izza2024distance}, our approach consistently achieves smaller explanation sizes, shorter generation times, and fewer $\checkValid$ calls. The improved explanation size results from our bound propagation-based traversal order, which yields more meaningful feature ranking than pixel-level saliency maps. Regarding generation time, their Dichotomic search (Algorithm 2 in \cite{izza2024distance}) restarts from scratch upon identifying each ``transition feature'' (i.e., explanatory feature), leading to redundant $\checkValid$ calls. In contrast, our binary search-inspired traversal avoids such inefficiencies.

For \cite{huang2023robustness}, their implementation of the QuickXplain algorithm~\cite{quickxplain} lacks the optimizations in our method.
There is also no evidence that any traversal order heuristic is employed. 
Accordingly, we first compare their results using a simple top-left to bottom-right order.  Unsurprisingly, we find that our method performs much better.
For a more fair comparison of their proposed algorithm, we compare again, but this time using our optimized bound propagation-based traversal order. We observe that with this better traversal order, their approach (Algorithm 2 in~\cite{huang2023robustness}) matches ours in terms of explanation sizes. 
 However, it consistently requires more $\checkValid$ calls and hence more time.

Earlier work by \cite{abduction} employs a toy network with a single hidden layer to distinguish between two MNIST digits (e.g., $1$ and $3$). However, their approach does not scale to the larger networks considered in our work and relies on naive traversal strategies, such as top-left to bottom-right or center-outward. As demonstrated in our comparison with Huang and Marques-Silva (2023) \cite{huang2023robustness}, these strategies fail to produce compact explanations relative to our method.

Finally, \cite{ore} focus on discrete perturbations, such as $k$-nearest neighbor manipulations in NLP models. Their setting is fundamentally different from ours, which considers continuous perturbations, rendering the two approaches incomparable.

\subsection{Comparison to Existing Non-formal Explanation Approaches}
\label{app:lime-anchors}

In Tables~\ref{tab:anchors-mnist} and \ref{tab:anchors-gtsrb}, we compare our approach with non-formal explainers LIME and Anchors—the latter incorporating a robustness perspective. We observe that our explanations are substantially smaller. The trade-off, however, is that generating formally provable explanations requires additional computation time.

\begin{figure*}[t]
    \centering
    \begin{subfigure}{0.49\linewidth}
        \centering
        \includegraphics[height=3cm]{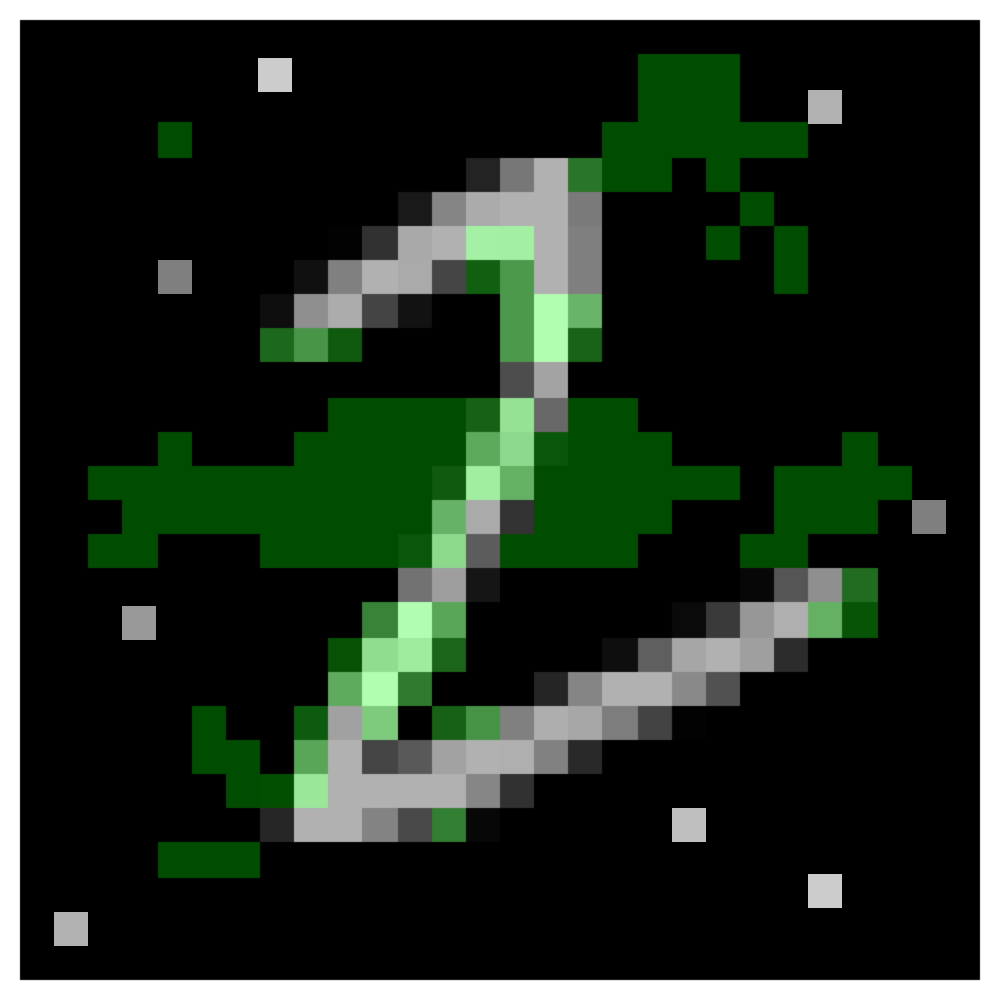}
        \caption{Pixel-level perturbations on the irrelevant features.}       
    \end{subfigure}
    \begin{subfigure}{0.49\linewidth}
        \centering
        \includegraphics[height=3cm]{figures/taxi/taxi-14-explanation.png}
        \caption{Brightness decreased by 5\%.}
    \end{subfigure}
    \caption{Examples of experiments investigating $\magnitude$-perturbations on the irrelevant features. (a): MNIST $\mathtt{2}$ with pixel-level perturbations that do not change prediction. (b): TaxiNet image with decreased brightness (e.g., to mimic different weather conditions) that maintains decision invariance.}
    \label{fig:perturbations}
\end{figure*}

We conducted a quantitative analysis of explanation robustness under the criterion ``robust w.r.t. \# perturbations'', using the same model architectures and standard benchmarks to ensure a fair comparison. Specifically, we generated 100 explanations per model and applied 10, 100, 500, and 1000 perturbations to each explanation by overlapping it with other images from the same dataset, following the protocol used in the original Anchors paper. We report the percentage of robust predictions and observe a sharp decline in robustness for Anchors (e.g., from $80\%$ to $7\%$). This analysis was not applied to LIME, as robustness is not its design objective. In contrast, our formal explanations guarantee decision invariance against any $\magnitude$-bounded perturbation. Figure~\ref{fig:perturbations} illustrates two example perturbations: pixel-level modifications to an MNIST digit “$\mathtt{2}$” and a brightness adjustment to a TaxiNet image, simulating plausible natural distortions such as varying weather conditions.

\subsection{Explanations from Adversarial Training}
\label{app:adv_training}

Studies have shown that neural networks can be ``fooled'' by adversarial inputs—test samples modified with small perturbations that are imperceptible to human eyes but cause the networks to misclassify them with high confidence. This vulnerability can be mitigated through adversarial training, which involves incorporating adversarial samples during the training process~\cite{goodfellow2014explaining}.

In Table~\ref{tab:adversarial}, we compare explanations generated from both normally trained and adversarially trained models.
To adversarially train the MNIST-FC model, after each training epoch, we applied the projected gradient descent (PGD) attack to generate adversarial examples from 50\% of randomly selected training samples, which were then combined with the remaining original training data. For testing, adversarial samples were crafted using PGD on a different MNIST network to prevent potential bias. The MNIST-FC model achieved a compromised accuracy of 23.66\% on this malicious test set, whereas the adversarially trained MNIST-FC-ADV model attained a significantly higher accuracy of 82.66\%.
We separated correctly and incorrectly predicted inputs, as adversarial samples tend to cause more misclassifications---often associated with larger explanation sizes (see Section~\ref{app:incorrect}).

\begin{figure*}[t]
    \centering
    \begin{subfigure}{\linewidth}
    \centering
        \includegraphics[width=0.15\linewidth]{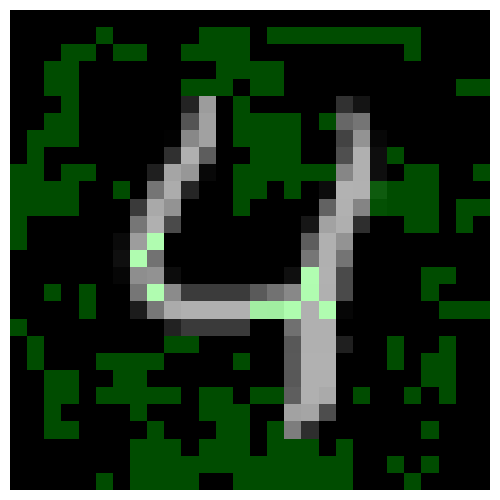}
        \includegraphics[width=0.15\linewidth]{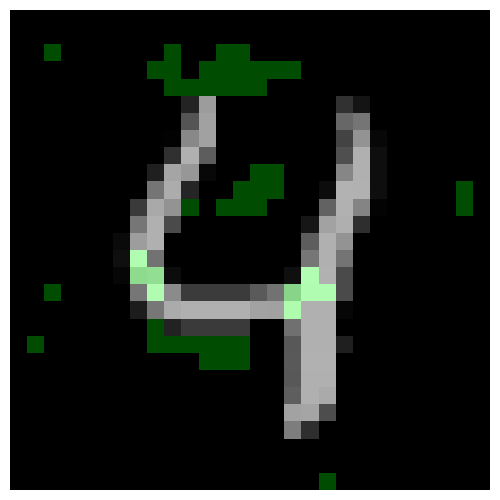}
        \quad \quad
        \includegraphics[width=0.15\linewidth]{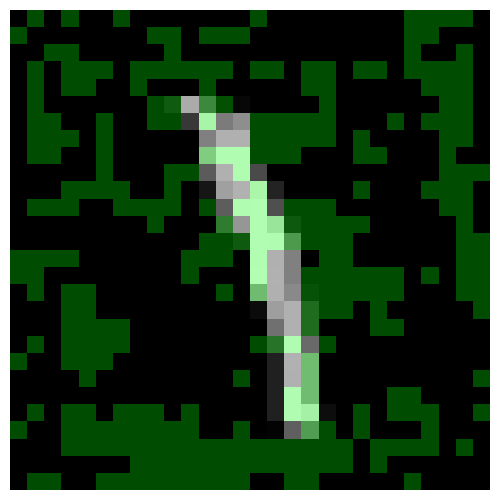}
        \includegraphics[width=0.15\linewidth]{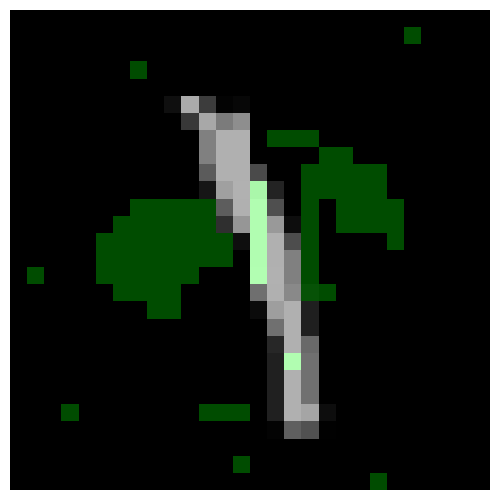}
        \caption{MNIST-FC and MNIST-FC-ADV on \emph{original} inputs.}
    \end{subfigure}
    \begin{subfigure}{\linewidth}
    \centering
        \includegraphics[width=0.15\linewidth]{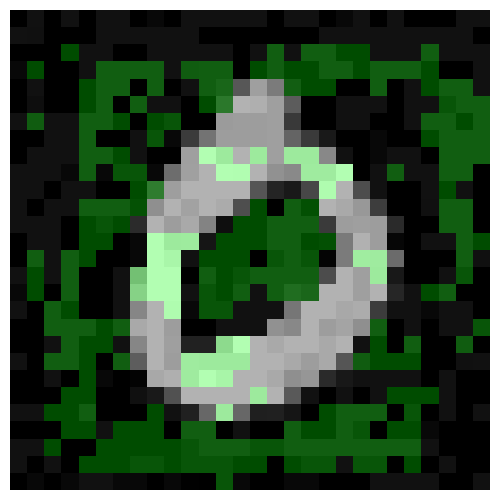}
        \includegraphics[width=0.15\linewidth]{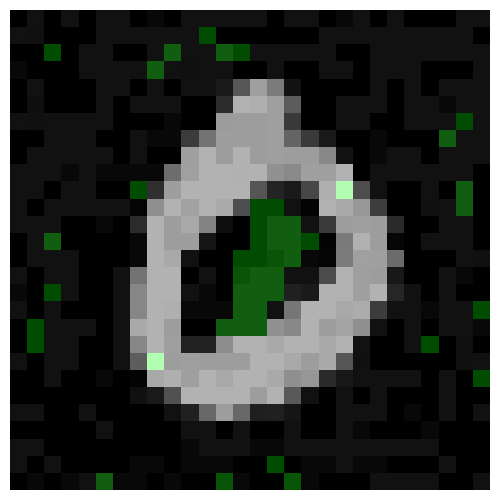}
        \quad \quad
        \includegraphics[width=0.15\linewidth]{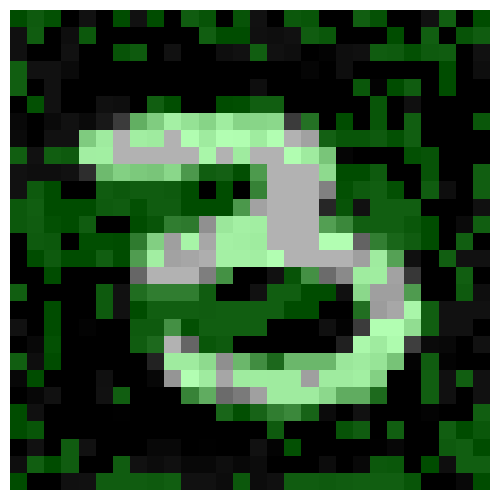}
        \includegraphics[width=0.15\linewidth]{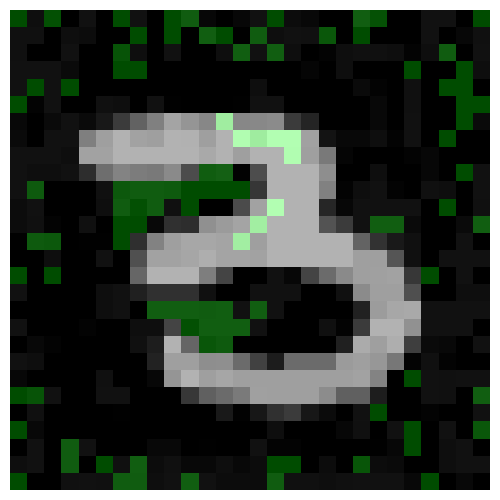}
        \caption{MNIST-FC and MNIST-FC-ADV on \emph{malicious} inputs.}
    \end{subfigure}
    \caption{Comparisons of explanations produced by normally trained MNIST-FC (left) and adversarially trained MNIST-FC-ADV (right) on both \emph{original} (top) and \emph{malicious} (bottom) inputs.}
    \label{fig:adv-examples}
\end{figure*}

Figure~\ref{fig:adv-examples} presents example explanations from the normally trained MNIST-FC and the adversarially trained MNIST-FC-ADV models on both original and malicious inputs. As shown, the adversarially trained model generates more compact explanations, whereas malicious inputs tend to produce larger ones. These observations are consistent with the quantitative results in Table~\ref{tab:adversarial}.

\subsection{Using Explanation Size to Detect Out-of-distribution Examples}
\label{app:ood}

\begin{figure*}[t]
    \centering
    \begin{subfigure}{0.49\linewidth}
        \centering
        \includegraphics[height=4.5cm]{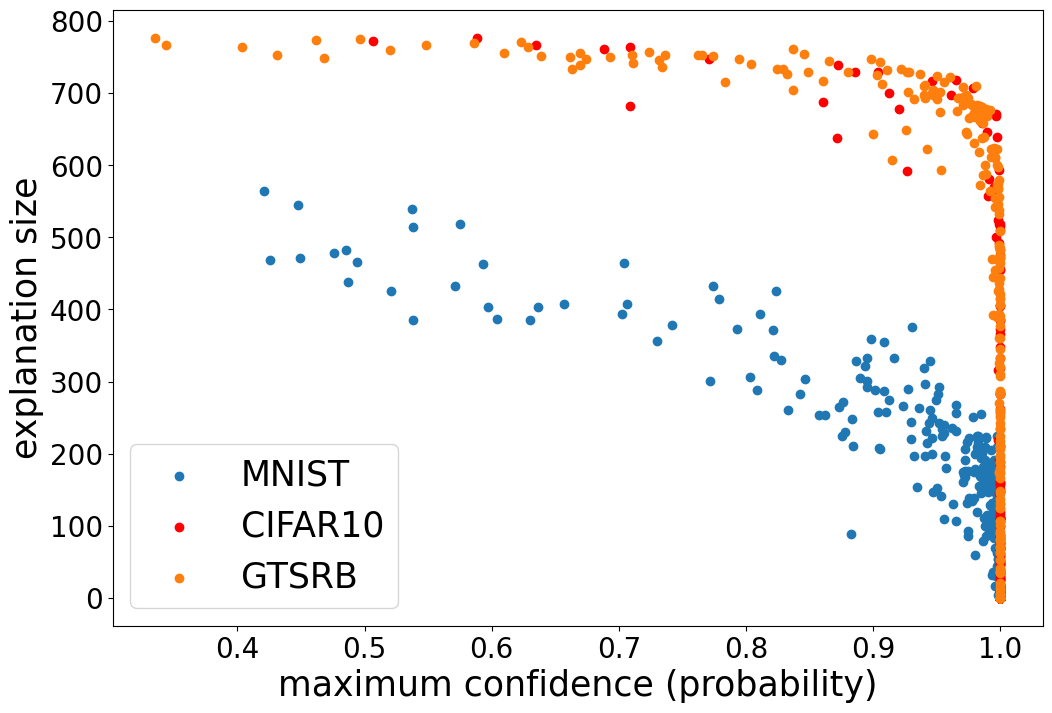}
        \caption{OOD: confidence \& size (\emph{linear} scale)}       
    \end{subfigure}
    \begin{subfigure}{0.49\linewidth}
        \centering
        \includegraphics[height=4.5cm]{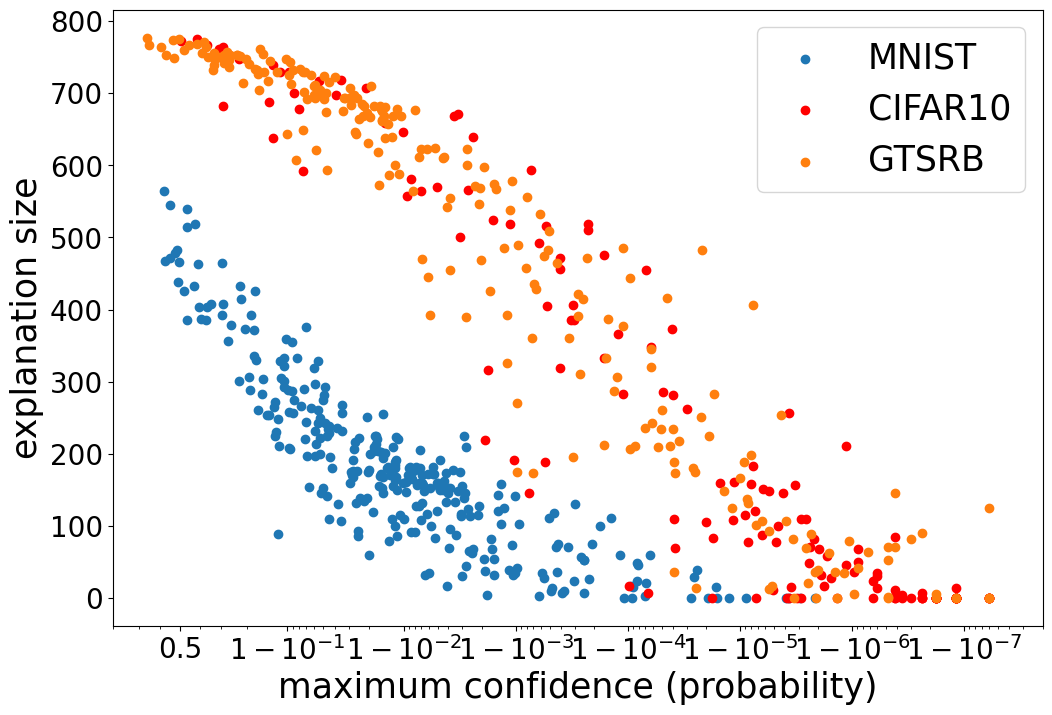}
        \caption{OOD: confidence \& size (\emph{log} scale)}       
    \end{subfigure}
    \begin{subfigure}{0.49\linewidth}
        \centering
        \includegraphics[height=3.8cm]{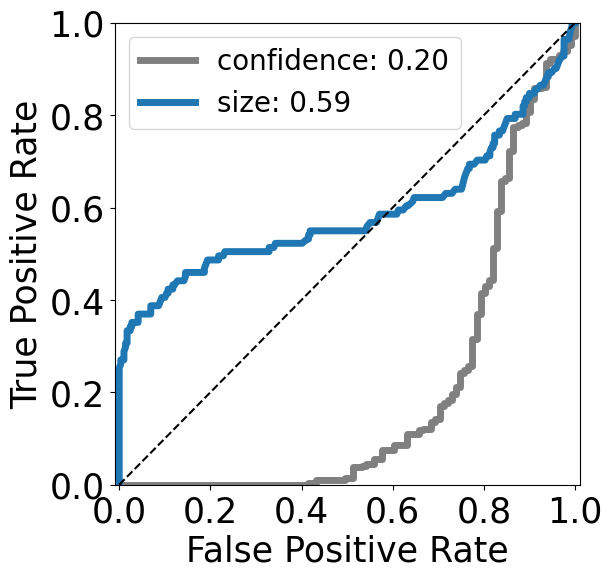}
        \caption{CIFAR10: AUROC}
    \end{subfigure}
    \begin{subfigure}{0.49\linewidth}
    \centering
        \includegraphics[height=3.8cm]{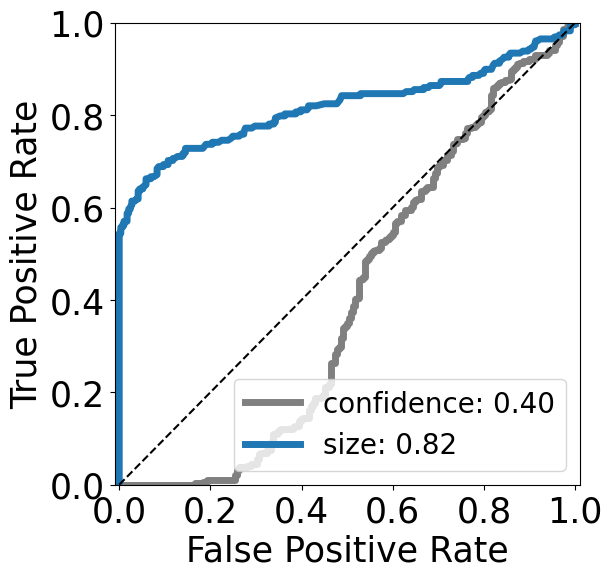}
        \caption{GTSRB: AUROC}
    \end{subfigure}
    \caption{Detecting \emph{out-of-distribution} examples from CIFAR10 and GTSRB for the MNIST-FC model. (a)(b) Maximum confidence and explanation size in \emph{linear} and \emph{log} scales. (c)(d) ROC curves and AUROC values for OOD samples from CIFAR10 and GTSRB, respectively.}
    \label{fig:ood-fc}
\end{figure*}

Deep neural networks can exhibit high-confidence predictions even when the test distribution differs from the training distribution, raising safety concerns~\cite{ood-confidence}. We explored the use of $\VeriXplus$ explanations as a method for detecting out-of-distribution (OOD) samples. Specifically, models were trained on the MNIST dataset, while GTSRB and CIFAR-10 images served as OOD inputs. We treated OOD samples as positive instances and evaluated performance using AUROC, comparing against the softmax probability baseline proposed by~\cite{softmax-baseline}. In addition to the MNIST-CNN results reported in Section~\ref{sec:ood}, we also conducted experiments on a fully connected MNIST-FC model. As shown in Figure~\ref{fig:ood-fc}, explanation size consistently outperforms the baseline in AUROC across both architectures, suggesting its effectiveness as an OOD indicator.
We emphasize that the goal of this section is to demonstrate the utility of formal explanations, rather than to outperform state-of-the-art OOD detection methods, which are beyond the scope of this paper.


\subsection{Using Explanation Size to Detect Misclassified Examples}
\label{app:incorrect}

Modern neural networks often exhibit overconfidence, producing incorrect predictions with confidence levels that exceed their actual accuracy~\cite{guo2017calibration,ood-confidence}. However, in safety-critical applications, it is essential for networks to flag potentially incorrect predictions for human oversight or intervention.

\begin{figure*}[t]
    \centering
    \begin{subfigure}{0.49\linewidth}
        \centering
        \includegraphics[height=4.5cm]{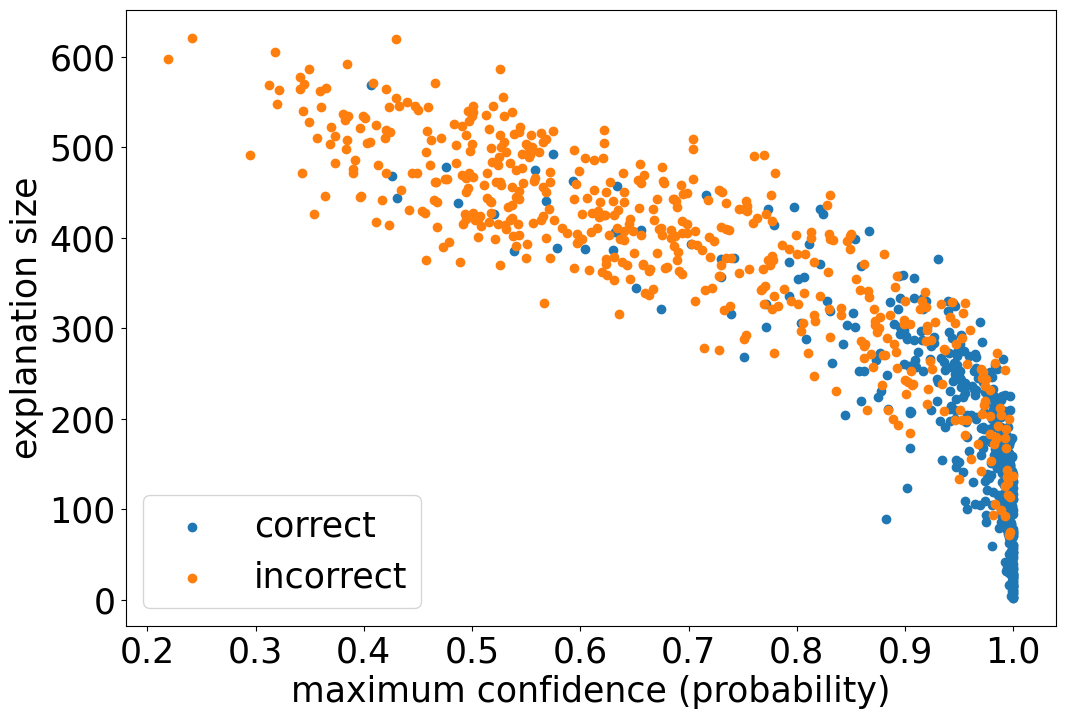}
        \caption{MNIST: confidence \& size (\emph{linear} scale)}       
        \label{fig:mnist_linear}
    \end{subfigure}
    \begin{subfigure}{0.49\linewidth}
        \centering
        \includegraphics[height=4.5cm]{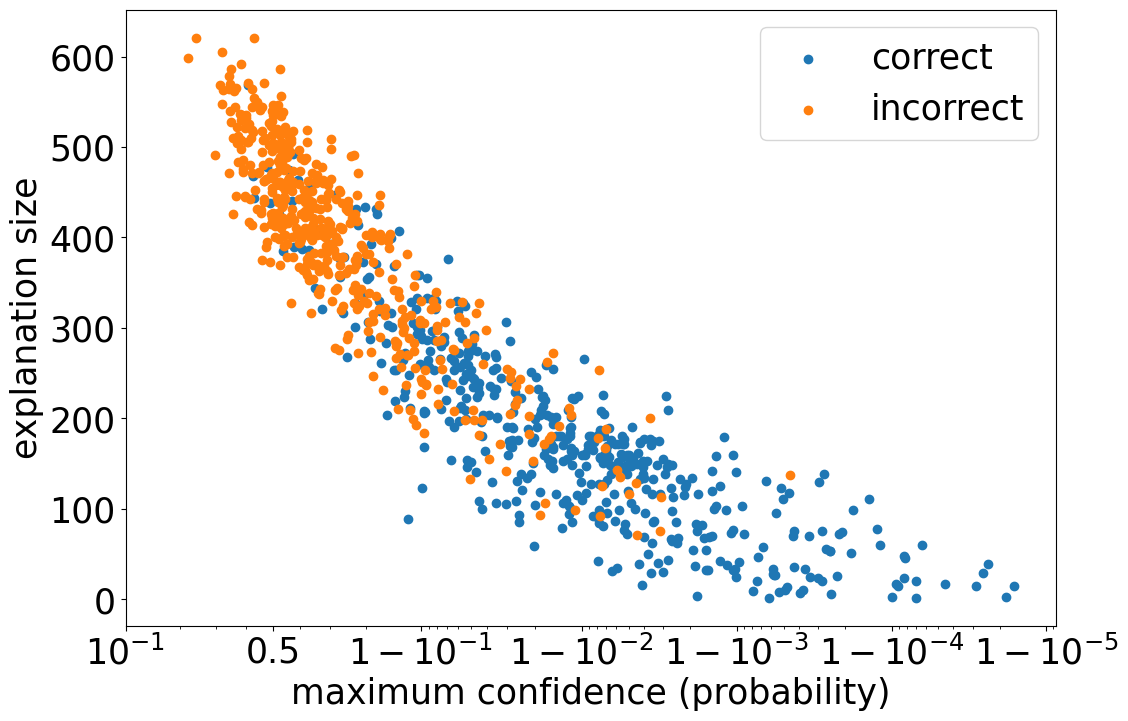}
        \caption{MNIST: confidence \& size (\emph{log} scale)}
        \label{fig:mnist_log}
    \end{subfigure}
    \caption{Maximum confidence and explanations size of \emph{correctly} and \emph{incorrectly} classified MNIST images in \emph{linear} and \emph{log} scales. As most probability values are in range $[0.9, 1]$, we also plot on \emph{log} scale for better visualization.}
    \label{fig:mnist_linear_log}
\end{figure*}

\begin{figure*}[t]
    \centering
        \begin{subfigure}[b]{0.49\linewidth}
            \centering
            \includegraphics[height=4cm]{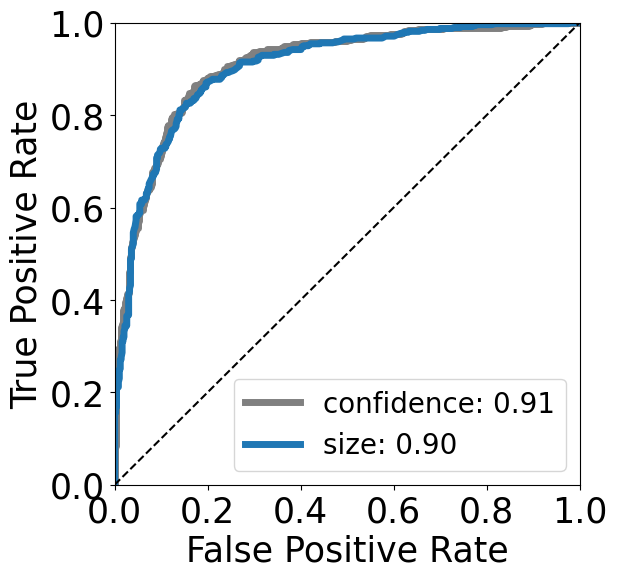}
            \vspace{4pt}
            \caption{MNIST: AUROC}
            \label{fig:incorrect_mnist_auroc}
        \end{subfigure}
        \begin{subfigure}[b]{0.49\linewidth}
            \centering
            \includegraphics[height=4.5cm]{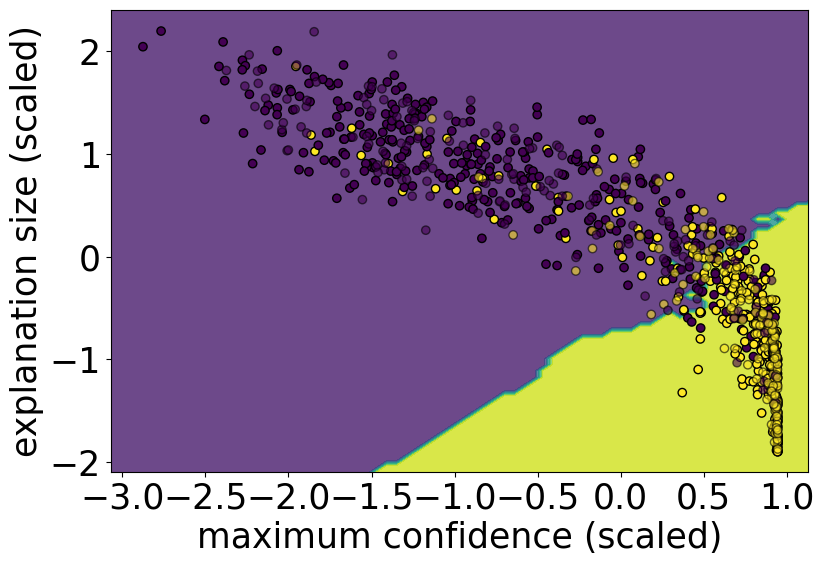}
            \caption{MNIST: A KNN classifier}
            \label{fig:incorrect_mnist_knn}
        \end{subfigure}
    \caption{Detecting \emph{incorrect} examples for MNIST.
    (a) ROC curves and AUROC values. (b) A KNN classifier.}
    \label{fig:incorrect_mnist}
\end{figure*}

\begin{figure*}[t]
    \centering
        \begin{subfigure}[b]{0.49\linewidth}
            \centering
            \includegraphics[height=4cm]{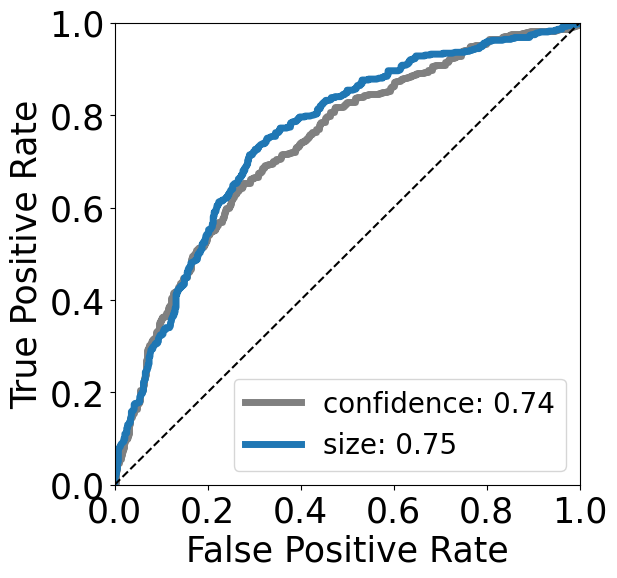}
            \vspace{4pt}
            \caption{GTSRB: AUROC}
            \label{fig:incorrect_gtsrb_auroc}
        \end{subfigure}
        \begin{subfigure}[b]{0.49\linewidth}
            \centering
            \includegraphics[height=4.5cm]{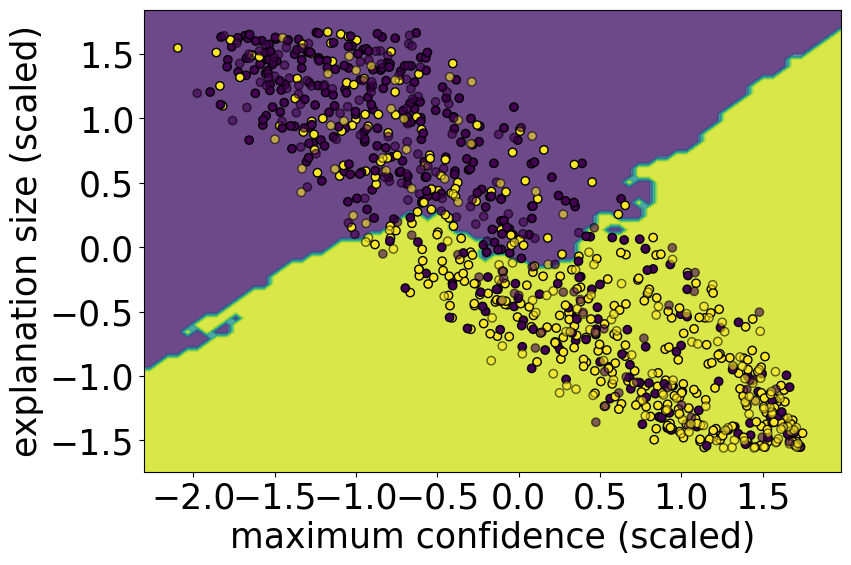}
            \caption{GTSRB: KNN classifier}
            \label{fig:incorrect_gtsrb_knn}
        \end{subfigure}

    \caption{Detecting \emph{incorrect} examples for GTSRB. 
    (a) ROC curves and AUROC values. (b) A KNN classifier.}
    \label{fig:incorrect_gtsrb}
\end{figure*}

We show that explanation size is a useful proxy for detecting \emph{incorrect} predictions. We collected $1000$ samples---$500$ correctly and $500$ incorrectly classified---for both MNIST and GTSRB, and present our analysis in Figures~\ref{fig:incorrect_mnist} and \ref{fig:incorrect_gtsrb}.
For MNIST, Figure~\ref{fig:mnist_linear_log} shows that there exists a significant separation between the correct and incorrect clusters.
Previously, \cite{softmax-baseline} proposed using the maximum softmax probabilities to detect erroneously classified examples. Following this, in Figures~\ref{fig:incorrect_mnist_auroc} and \ref{fig:incorrect_gtsrb_auroc}, we plot the receiver operating characteristic (ROC) curves and compute the AUROC (area under the ROC curve) values when using maximum confidence (gray) and explanation size (blue) independently. We observe that our explanation size has a competitive effect on both datasets. 

\begin{figure*}
    \centering
    \begin{subfigure}{0.8\linewidth}
        \centering
        \includegraphics[width=\linewidth]{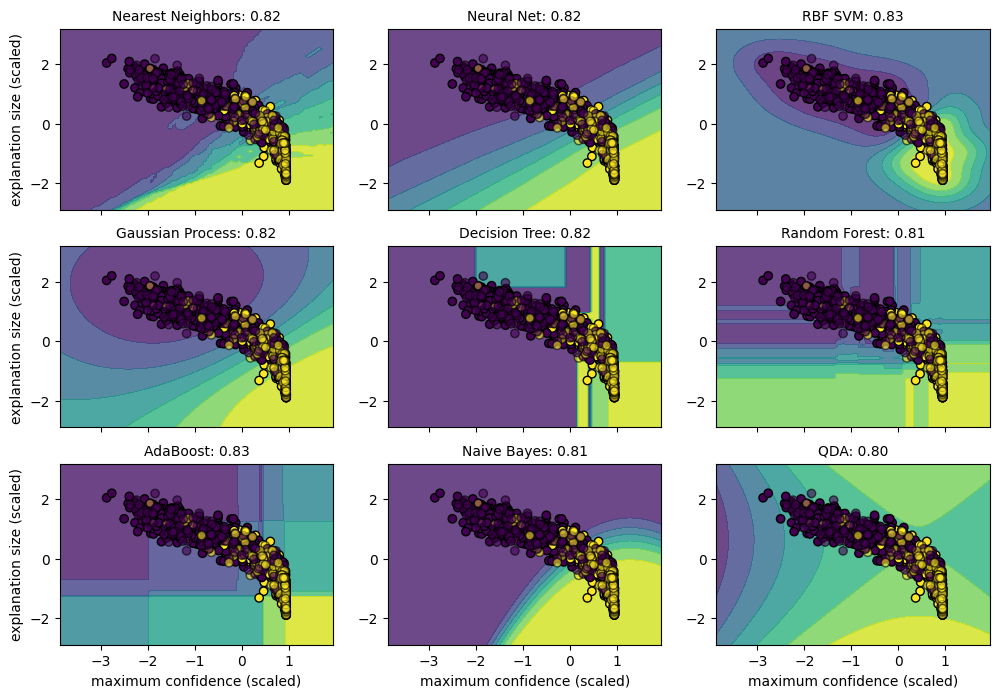}
        \caption{MNIST}
    \end{subfigure}
    \begin{subfigure}{0.8\linewidth}
        \centering
        \includegraphics[width=\linewidth]{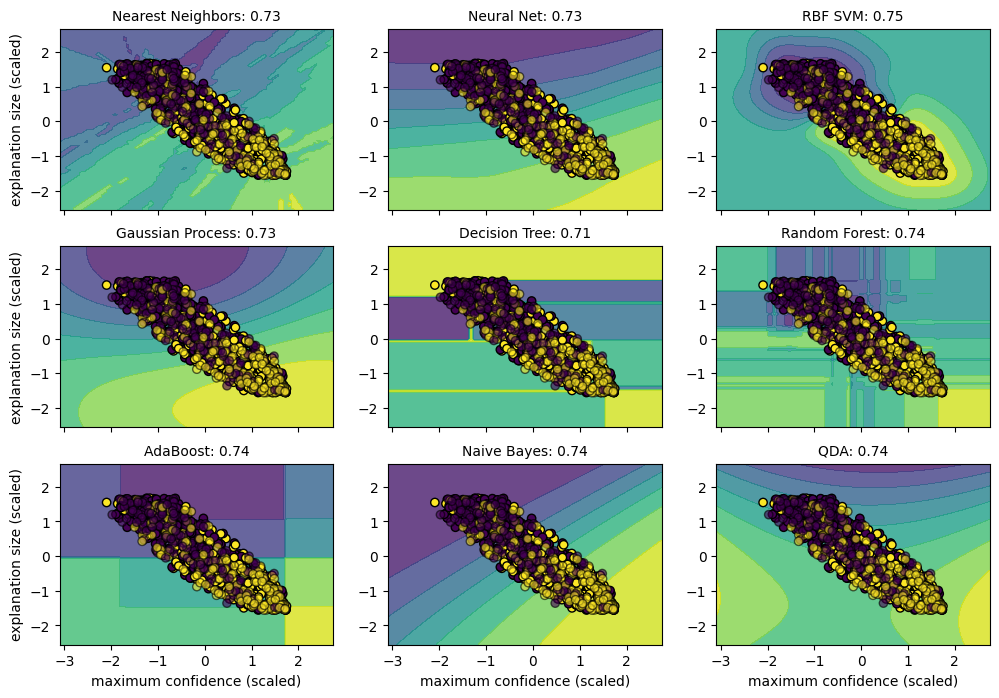}
        \caption{GTSRB}
    \end{subfigure}
    \caption{Various classifiers using maximum confidence and explanation size to detect \emph{correct} and \emph{incorrect} examples in the MNIST (top) and GTSRB (bottom) datasets. Accuracy of each classifier is plotted above each sub-figure, e.g., Radial Basis Function Support Vector Machine (RBF SVM) achieves $83\%$ and $75\%$, respectively.}
    \label{fig:classifiers}
\end{figure*}

Furthermore, utilizing both confidence and size, we trained classifiers to detect whether the prediction of an unseen sample is correct or erroneous. We split the $1000$ samples into a training set ($700$ samples) and a test set ($300$ samples), and then trained a $k$-nearest neighbor (KNN) classifier for each dataset. As in Figures~\ref{fig:incorrect_mnist_knn} and \ref{fig:incorrect_gtsrb_knn}, these two classifiers achieve $82\%$ and $73\%$ accuracy, respectively. Without loss of generosity, in Figure~\ref{fig:classifiers}, we also trained various other classifiers, including neural networks, support vector machines (SVMs), and decision trees, and observe that they achieve similar accuracies. 


\begin{table*}[t]
\caption{Experimental evidence for the visualization in Figure~\ref{fig:verix_verix+}, where explanations from $\VeriX$ and $\VeriXplus$ are compared on the MNIST, GTSRB, and CIFAR10 datasets. Here we present the size and time measurements, and calculate their corresponding reduction rate in percentage, e.g., for GTSRB-CNN, $(569.0-355.3)/569.0 = 37.56\%$. }
\label{tab:size-reduction}
    \centering
    \setlength{\tabcolsep}{4pt} 
    \renewcommand{\arraystretch}{1.2}
\begin{tabular}{l|cc|cc|c|c}
\toprule
                                      & \multicolumn{2}{c|}{$\VeriX$}      & \multicolumn{2}{c|}{$\VeriXplus$}            & \multirow{4}{*}{size reduction} & \multirow{4}{*}{time reduction} \\
\multicolumn{1}{c|}{traversal order}  & \multicolumn{2}{c|}{saliency}  & \multicolumn{2}{c|}{bound propagation} &                                 &                                 \\
\multicolumn{1}{c|}{traversal procedure} & \multicolumn{2}{c|}{sequential} & \multicolumn{2}{c|}{QuickXplain}       &                                 &                                 \\
\multicolumn{1}{c|}{metrics}          & size           & time           & size               & time              &                                 &                                 \\ \hline \hline
MNIST-FC                              & $186.7$          & $92.3$           & $179.0$              & $32.2$              & $\mathbf{4.12\%}$                & $\mathbf{65.11\%}$                \\
MNIST-CNN                             & $105.7$          & $439.2$          & $100.6$              & $47.4$              & $\mathbf{4.82\%}$                & $\mathbf{89.21\%}$               \\
GTSRB-FC                              & $529.4$          & $614.9$          & $333.4$             & $196.1$             & $\mathbf{37.02\%}$               & $\mathbf{68.11\%}$               \\
GTSRB-CNN                             & $569.0$          & $1897.6$         & $355.3$              & $330.1$             & $\mathbf{37.56\%}$               & $\mathbf{82.60\%}$               \\
CIFAR10-FC                            & $588.9$          & $438.0$          & $465.7$              & $316.9$             & $\mathbf{20.92\%}$               & $\mathbf{27.65\%}$               \\
CIFAR10-CNN                           & $664.8$          & $1617.0$         & $552.8$              & $482.2$             & $\mathbf{16.85\%}$               & $\mathbf{70.18\%}$               \\ \bottomrule
\end{tabular}
\end{table*}

\begin{table*}[t]
\centering
    \setlength{\tabcolsep}{4pt} 
\renewcommand{\arraystretch}{1.2}
\caption{$\VeriXplus$ vs. Anchors vs. LIME on the MNIST dataset, showing average explanation size (number of pixels), generation time (seconds), and a robustness evaluation of the produced explanations against perturbations.}
\label{tab:anchors-mnist}
    \begin{tabular}{c|cc|cccc|cc|cccc}
        \toprule
        \multicolumn{1}{c|}{\multirow{2}{*}{}} & \multicolumn{6}{c|}{MNIST-CF} & \multicolumn{6}{c}{MNIST-CNN} \\
        \multicolumn{1}{c|}{} & size & time & \multicolumn{4}{c|}{robust wrt \# perturbations} & size & time & \multicolumn{4}{c}{robust wrt \# perturbations} \\ \midrule \midrule
        \multirow{1}{*}{$\VeriXplus$} & $\bf{179.5}$ & $\bf{28.4}$ & \multicolumn{4}{c|}{$\bf{100\%}$} & $\bf{100.8}$ & $\bf{42.0}$ & \multicolumn{4}{c}{$\bf{100\%}$} \\ \hline
        \multicolumn{1}{c|}{\multirow{2}{*}{Anchors}} & \multirow{2}{*}{$532.6$} & \multirow{2}{*}{$9.1$} & $10$ & $100$ & $500$ & $1000$ & \multirow{2}{*}{$520.6$} & \multirow{2}{*}{$8.8$} & $10$ & $100$ & $500$ & $1000$ \\
        \multicolumn{1}{c|}{} & & & $80\%$ & $24\%$ & $7\%$ & $7\%$ & & & $84\%$ & $28\%$ & $17\%$ & $11\%$ \\ \hline
        \multicolumn{1}{c|}{LIME} & $428.6$ & $8.0$ & \multicolumn{4}{c|}{--} & $419.8$ & $8.1$ & \multicolumn{4}{c}{--} \\
        \bottomrule        
    \end{tabular}
\end{table*}

\begin{table*}[t]
\centering
    \setlength{\tabcolsep}{4pt} 
\renewcommand{\arraystretch}{1.2}
\caption{$\VeriXplus$ vs. Anchors vs. LIME on the GTSRB dataset, showing average explanation size (number of pixels), generation time (seconds), and a robustness evaluation of the produced explanations against perturbations.}
\label{tab:anchors-gtsrb}
    \begin{tabular}{c|cc|cccc|cc|cccc}
        \toprule
        \multicolumn{1}{c|}{\multirow{2}{*}{}} & \multicolumn{6}{c|}{GTSRB-CF} & \multicolumn{6}{c}{GTSRB-CNN} \\
        \multicolumn{1}{c|}{} & size & time & \multicolumn{4}{c|}{robust wrt \# perturbations} & size & time & \multicolumn{4}{c}{robust wrt \# perturbations} \\ \midrule \midrule
        \multirow{1}{*}{$\VeriXplus$} & $\bf{333.6}$ & $\bf{149.7}$ & \multicolumn{4}{c|}{$\bf{100\%}$} & $\bf{355.8}$ & $\bf{251.0}$ & \multicolumn{4}{c}{$\bf{100\%}$} \\ \hline
        \multicolumn{1}{c|}{\multirow{2}{*}{Anchors}} & \multirow{2}{*}{$692.0$} & \multirow{2}{*}{$32.2$} & $10$ & $100$ & $500$ & $1000$ & \multirow{2}{*}{$576.5$} & \multirow{2}{*}{$5.9$} & $10$ & $100$ & $500$ & $1000$ \\
        \multicolumn{1}{c|}{} & & & $74\%$ & $38\%$ & $30\%$ & $30\%$ & & & $73\%$ & $26\%$ & $16\%$ & $14\%$ \\ \hline
        \multicolumn{1}{c|}{LIME} & $531.9$ & $8.0$ & \multicolumn{4}{c|}{--} & $516.7$ & $6.9$ & \multicolumn{4}{c}{--} \\
        \bottomrule        
    \end{tabular}
\end{table*}

%


\begin{figure*}[t]
    \centering
    \begin{subfigure}{0.49\linewidth}
        \centering
        \includegraphics[height=4.5cm]{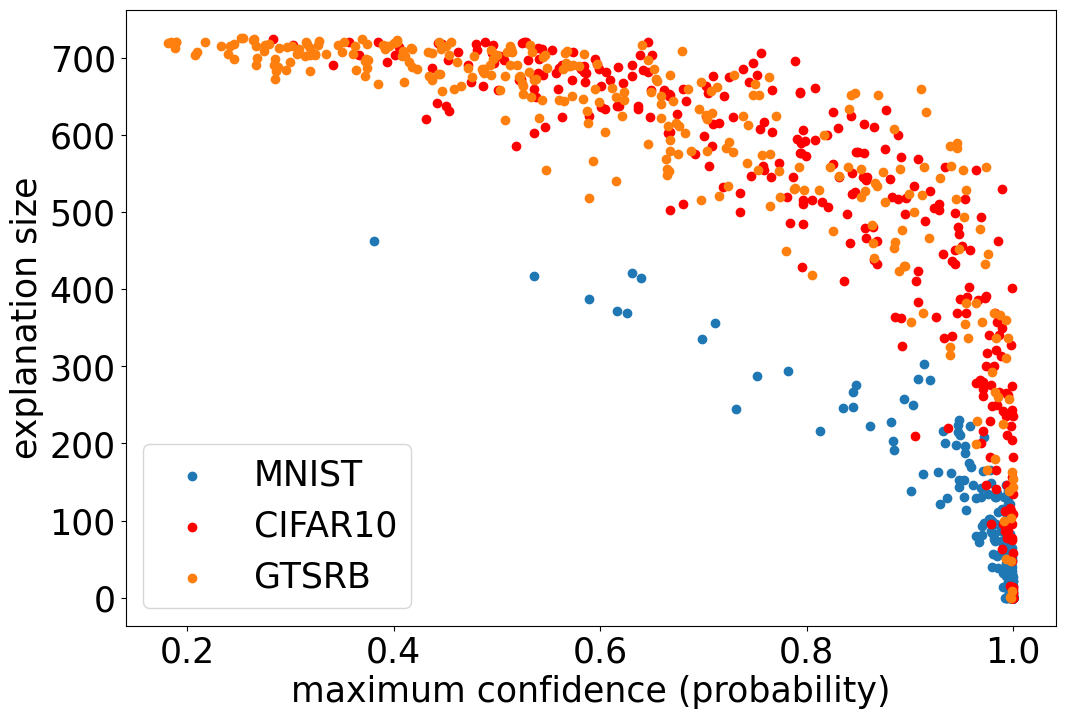}
        \caption{OOD: confidence \& size (\emph{linear} scale)}       
        \label{fig:ood_cnn_linear}
    \end{subfigure}
    \begin{subfigure}{0.49\linewidth}
        \centering
        \includegraphics[height=4.5cm]{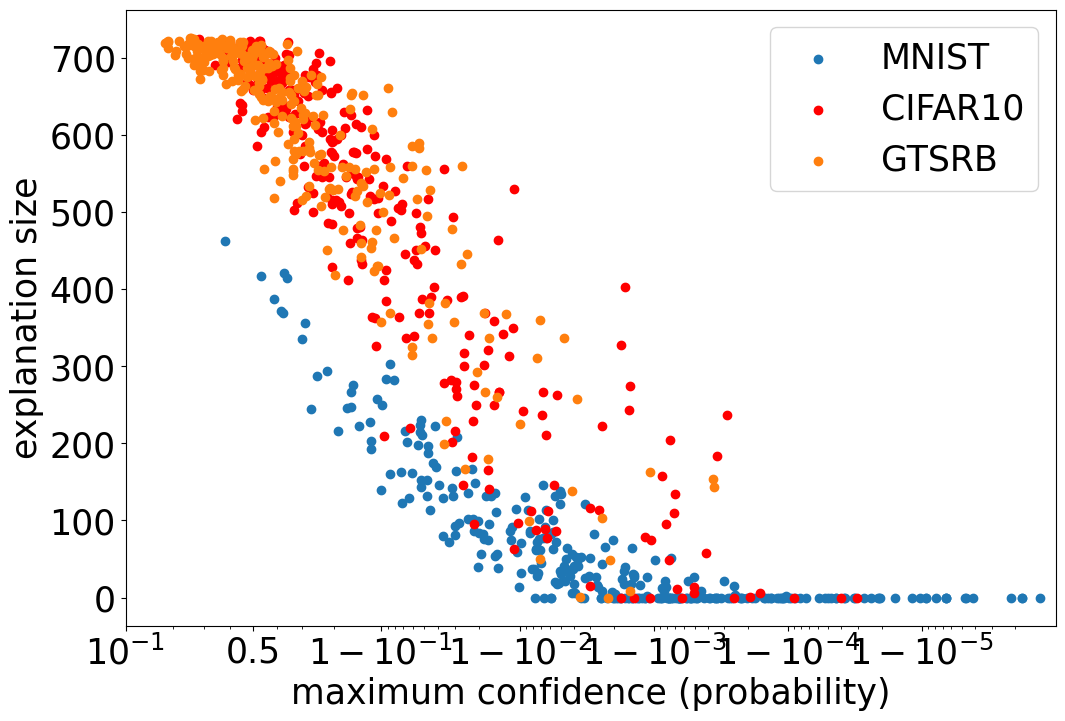}
        \caption{OOD: confidence \& size (\emph{log} scale)}
        \label{fig:ood_cnn_log}
    \end{subfigure}
    \caption{Maximum confidence and explanation size in \emph{linear} and \emph{log} scales of \emph{in-distribution} (MNIST) and \emph{out-of-distribution} (CIFAR10 and GTSRB) images that are classified by the MNIST-CNN model.}
    \label{fig:ood_cnn_linear_log}
\end{figure*}
%
%

\newpage
\clearpage
\twocolumn

\section{Model Specifications}
\label{app:models}

We evaluated our framework on standard image benchmarks including the MNIST~\cite{mnist}, GTSRB~\cite{gtsrb}, and CIFAR10~\cite{cifar10} datasets, and trained both fully connected and convolutional models. Note that, when calling the $\marabou$~\cite{marabou,marabou2} tool to perform the $\solver$ functionality, in order for the queries to be \emph{decidable}, the outputs of the networks need to be raw logits before the final $\mathsf{softmax}$ activation. For this, one needs to specify the $\mathtt{outputName}$ argument of the $\mathtt{read\_onnx}$ function to be the pre-$\mathsf{softmax}$ logits. We remark that this does not change the decisions of a network.
As a workaround for this, one can also train the model by setting $\mathtt{from\_logits}$=$\true$ in the loss function.

\subsection{MNIST, GTSRB, and CIFAR10}

The MNIST dataset consists of grayscale handwritten digits with resolution $28 \times 28 \times 1$. The architectures for the fully connected and convolutional models trained on MNIST are shown in Tables~\ref{models:mnist-10x2} and~\ref{models:mnist-cnn}, achieving accuracies of $93.76\%$ and $96.29\%$, respectively.

The GTSRB dataset comprises color traffic sign images of size $32 \times 32 \times 3$. To improve accuracy, we selected the 10 most frequent classes from the original 43. The corresponding fully connected and convolutional model architectures are given in Tables~\ref{models:gtsrb-10x2} and~\ref{models:gtsrb-cnn}, with resulting accuracies of $85.93\%$ and $90.32\%$.

The CIFAR10 dataset also contains $32 \times 32 \times 3$ images across 10 categories. The architectures used for this dataset are provided in Tables~\ref{models:cifar10-10x2} and~\ref{models:cifar10-cnn}.

\subsection{Transformers}

We adapt the transformer architecture proposed in \cite{attention} for two tasks: estimating aircraft cross-track distance, using TaxiNet-Transformer (Table~\ref{tab:transformer-taxi}), and performing sentiment analysis on the IMDB dataset, using IMDB-Transformer (Table~\ref{tab:transformer-imdb}). The core of each model is the transformer block, implemented via a $\mathtt{MultiHeadAttention}$ layer.

\newpage

\begin{table}[h!]
    \caption{Architecture of the MNIST-FC model.}
    \label{models:mnist-10x2}
    \centering
\resizebox{\linewidth}{!}{
    \begin{tabular}{c|c|c|c}
    \toprule
    Layer Type &  Input Shape & Output Shape & Activation \\
    \midrule
    Flatten & 28 $\cross$ 28 $\cross$ 1 & 784 & -- \\
    Fully Connected & 784 & 10 & $\mathsf{ReLU}$ \\
    Fully Connected & 10 & 10 & $\mathsf{ReLU}$ \\
    Output & 10 & 10 & -- \\
    \bottomrule
    \end{tabular}
    }
\end{table}

\begin{table}[h!]
    \caption{Architecture of the MNIST-CNN model.}
    \label{models:mnist-cnn}
    \centering
\resizebox{\linewidth}{!}{
    \begin{tabular}{c|c|c|c}
    \toprule
    Layer Type &  Input Shape & Output Shape & Activation \\
    \midrule
    Convolution 2D& $28 \cross 28 \cross 1$ & $13 \cross 13 \cross 4$ & -- \\
    Convolution 2D& $13 \cross 13 \cross 4$ & $6 \cross 6 \cross 4$ & -- \\
    Flatten & $6 \cross 6 \cross 4$ & 144 & -- \\
    Fully Connected & 144 & 20 & $\mathsf{ReLU}$ \\
    Output & 20 & 10 & -- \\
    \bottomrule
    \end{tabular}
    }
\end{table}

\begin{table}[h!]
    \caption{Architecture of the GTSRB-FC model.}
    \label{models:gtsrb-10x2}
    \centering
\resizebox{\linewidth}{!}{
    \begin{tabular}{c|c|c|c}
    \toprule
    Layer Type &  Input Shape & Output Shape & Activation \\
    \midrule
    Flatten & 32 $\cross$ 32 $\cross$ 3 & 3072 & -- \\
    Fully Connected & 3072 & 10 & $\mathsf{ReLU}$ \\
    Fully Connected & 10 & 10 & $\mathsf{ReLU}$ \\
    Output & 10 & 10 & -- \\
    \bottomrule
    \end{tabular}
    }
\end{table}

\begin{table}[h!]
    \caption{Architecture of the GTSRB-CNN model.}
    \label{models:gtsrb-cnn}
    \centering
\resizebox{\linewidth}{!}{
    \begin{tabular}{c|c|c|c}
    \toprule
    Layer Type &  Input Shape & Output Shape & Activation \\
    \midrule
    Convolution 2D& $32 \cross 32 \cross 3$ & $15 \cross 15 \cross 4$ & -- \\
    Convolution 2D& $15 \cross 15 \cross 4$ & $7 \cross 7 \cross 4$ & -- \\
    Flatten & $7 \cross 7 \cross 4$ & 196 & -- \\
    Fully Connected & 196 & 20 & $\mathsf{ReLU}$ \\
    Output & 20 & 10 & -- \\
    \bottomrule
    \end{tabular}
    }
\end{table}


\begin{table}[h!]
    \caption{Architecture of the CIFAR10-FC model.}
    \label{models:cifar10-10x2}
    \centering
\resizebox{\linewidth}{!}{
    \begin{tabular}{c|c|c|c}
    \toprule
    Layer Type &  Input Shape & Output Shape & Activation \\
    \midrule
    Flatten & 32 $\cross$ 32 $\cross$ 3 & 3072 & -- \\
    Fully Connected & 3072 & 10 & $\mathsf{ReLU}$ \\
    Fully Connected & 10 & 10 & $\mathsf{ReLU}$ \\
    Output & 10 & 10 & -- \\
    \bottomrule
    \end{tabular}
    }
\end{table}

\begin{table}[h!]
    \caption{Architecture of the CIFAR10-CNN model.}
    \label{models:cifar10-cnn}
    \centering
\resizebox{\linewidth}{!}{
    \begin{tabular}{c|c|c|c}
    \toprule
    Layer Type &  Input Shape & Output Shape & Activation \\
    \midrule
    Convolution 2D& $32 \cross 32 \cross 3$ & $15 \cross 15 \cross 4$ & -- \\
    Convolution 2D& $15 \cross 15 \cross 4$ & $7 \cross 7 \cross 4$ & -- \\
    Flatten & $7 \cross 7 \cross 4$ & 196 & -- \\
    Fully Connected & 196 & 20 & $\mathsf{ReLU}$ \\
    Output & 20 & 10 & -- \\
    \bottomrule
    \end{tabular}
    }
\end{table}

\begin{table*}[h!]
    \caption{Architecture for the TaxiNet-Transformer model.} 
    \label{tab:transformer-taxi}
    \centering		
    \renewcommand{\arraystretch}{1.2}
    \begin{tabular}{c|c|c}
        \toprule
        Layer Type & Parameter Size & Activation \\
        \midrule
        Input & $\mathtt{shape}$=$(27, 54, 1$) & -- \\
        Flatten &  -- & -- \\
        Fully Connected & $16$, $\mathtt{kernel\_initializer}$=``$\mathtt{he\_uniform}$'' & $\mathsf{ReLU}$ \\
        Reshape & $(2, 8)$ & -- \\
        MultiHeadAttention & $\mathtt{num\_heads}$=$2$, $\mathtt{key\_dim}$=$4$ & -- \\
        Flatten &  -- & -- \\
        Fully Connected & $1$ & -- \\
        \bottomrule
    \end{tabular}
\end{table*}

\begin{table*}[h!]
    \caption{Architecture for the IMDB-Transformer model.} 
    \label{tab:transformer-imdb}
    \centering		
    \renewcommand{\arraystretch}{1.2}
    \begin{tabular}{c|c|c}
        \toprule
        Layer Type & Parameter Size & Activation \\
        \midrule
        Input & $\mathtt{shape}$=$(1000,)$ & -- \\
        Embedding & $\mathtt{vocab\_size}$=$10000$, $\mathtt{embedding\_dim}$=$4$, $\mathtt{input\_length}$=$1000$ & -- \\
        Flatten &  -- & -- \\
        Fully Connected & $16$ & $\mathsf{ReLU}$ \\
        Reshape & $(2, 8)$ & -- \\
        MultiHeadAttention & $\mathtt{num\_heads}$=$2$, $\mathtt{key\_dim}$=$4$ & -- \\
        Flatten &  -- & -- \\
        Fully Connected & $1$ & $\mathsf{Sigmoid}$ \\
        \bottomrule
    \end{tabular}
\end{table*}




\end{document}